\documentclass[twoside]{article} % For LaTeX2e
\usepackage{iclr2026_conference,times}

\usepackage{amsmath,amsfonts,bm}

\def\eqref#1{equation~\ref{#1}}
\def\1{\bm{1}}

\DeclareMathAlphabet{\mathsfit}{\encodingdefault}{\sfdefault}{m}{sl}
\SetMathAlphabet{\mathsfit}{bold}{\encodingdefault}{\sfdefault}{bx}{n}

\usepackage{hyperref}
\usepackage{url}
\usepackage{graphicx}
\usepackage{amsmath}
\usepackage{amssymb}
\usepackage{booktabs}
\usepackage{float}
\usepackage{mathtools}
\usepackage{wrapfig}
\usepackage{subcaption}

\title{Efficient Sliced Wasserstein Distance \\ Computation via Adaptive Bayesian \\ Optimization}

\author{Manish Acharya and David Hyde \\
Department of Computer Science\\
Vanderbilt University\\
Nashville, TN , USA \\
\texttt{\{manish.acharya,david.hyde.1\}@vanderbilt.edu}
}

\iclrfinalcopy % Uncomment for camera-ready version, but NOT for submission.
\begin{document}

\maketitle
\thispagestyle{empty}

\begin{abstract}
The sliced Wasserstein distance (SW) reduces optimal transport on $\mathbb{R}^d$ to a sum of one-dimensional projections, and thanks to this efficiency, it is widely used in geometry, generative modeling, and registration tasks.
Recent work shows that quasi-Monte Carlo constructions for computing SW (QSW) yield direction sets with excellent approximation error.
This paper presents an alternate, novel approach: learning directions with Bayesian optimization (BO), particularly in settings where SW appears inside an optimization loop (e.g., gradient flows).
We introduce a family of drop-in selectors for projection directions: \textbf{BOSW}, a one-shot BO scheme on the unit sphere; \textbf{RBOSW}, a periodic-refresh variant; \textbf{ABOSW}, an adaptive hybrid that seeds from competitive QSW sets and performs a few lightweight BO refinements; and \textbf{ARBOSW}, a restarted hybrid that periodically relearns directions during optimization.
Our BO approaches can be composed with QSW and its variants (demonstrated by ABOSW/ARBOSW) and require no changes to downstream losses or gradients.
We provide numerical experiments where our methods achieve state-of-the-art performance, and on the experimental suite of the original QSW paper, we find that ABOSW and ARBOSW can achieve convergence comparable to the best QSW variants with modest runtime overhead.
%We release code with fixed seeds and configurations to support faithful replication (see supplementary material).
\end{abstract}

\section{Introduction}
\label{sec:intro}

Optimal transport (OT) is a mathematical framework for measuring distances between probability measures \citep{villani2008optimal}.
OT has seen increasing interest from the learning community in recent years since, for example, collections of sample data can be interpreted as empirical distributions.
In particular, the Wasserstein distance is a popular metric for OT tasks in the learning literature (compared to, e.g., Kullback-Lieibler (KL) divergence, which is not symmetric, nor does it satisfy the triangle inequality) \citep{solomon2014wasserstein,montavon2016wasserstein,kolouri2017optimal}.

However, evaluating the conventional Wasserstein distance (WD) is computationally prohibitive ($\mathcal{O}(n^3 \log n)$ in time and $\mathcal{O}(n^2)$ in space \citep{nguyen2023}), particularly for higher-dimensional measures, where computing Wasserstein distances can become the bottleneck of an application \citep{kolouri2019generalized}.
The sliced Wasserstein (SW) distance \citep{bonneel2015} offers a scalable alternative to the classical WD by averaging one-dimensional Wasserstein costs over projections on the unit sphere, giving $\mathcal{O}(n \log n)$ time and $\mathcal{O}(n)$ memory for discrete measures via sorting, and enabling its use across geometry, imaging, and generative modeling tasks~\citep{peyre2019computational,bonneel2015}\footnote{Note that \citet{bonnotte2013unidimensional} showed SW is equivalent to WD for compactly supported measures.}.
Formally, the SW between $\mu,\nu \in \mathcal{P}_p(\mathbb{R}^d)$ ($p \geq 1$) is
\begin{equation}
\mathrm{SW}_p^p(\mu,\nu) = \mathbb{E}_{\theta \sim \mathcal{U}(\mathbb{S}^{d-1})} \left[ W_p^p(\theta_\sharp \mu, \theta_\sharp \nu) \right] .
\label{eq:sw-continuous}
\end{equation}
%\noindent\textbf{Notation for Eq.~(1).}
Here, $d$ is the ambient dimension, and $p\ge 1$ is the order of the Wasserstein cost.
$\mathcal{P}_p(\mathbb{R}^d)$ denotes the set of Borel probability measures on $\mathbb{R}^d$
with finite $p$-th moment, i.e., $\int \|x\|^p\,\mathrm{d}\mu(x)<\infty$.
$\mathbb{S}^{d-1}$ is the unit sphere, and $\mathcal{U}(\mathbb{S}^{d-1})$ is the uniform distribution on it.
For a direction $\theta\in\mathbb{S}^{d-1}$, the pushforward $\theta_\sharp\mu$ is the distribution of the scalar projection $\theta^\top X$ when $X\!\sim\!\mu$ (i.e., under $x\mapsto \theta^\top x$).
$W_p(\cdot,\cdot)$ is the $p$-Wasserstein distance on $\mathbb{R}$; we average its $p$-th power over directions, i.e.,
$\mathrm{SW}_p^p(\mu,\nu)=\mathbb{E}_{\theta}[W_p^p(\theta_\sharp\mu,\theta_\sharp\nu)]$.
Equivalently, one may write $\mathrm{SW}_p(\mu,\nu)=(\mathbb{E}_{\theta}[W_p^p(\theta_\sharp\mu,\theta_\sharp\nu)])^{1/p}$.

%\paragraph{Finite-direction estimator and objective.}
Of course, in practice, the expectation---which integrates over all directions---must be estimated using a finite set of directions.
Given $L$ directions (``slices'') $\Theta_L=\{\theta_\ell\}_{\ell=1}^L$, we estimate
\begin{equation}
\widehat{\mathrm{SW}}_p^p(\mu,\nu;\Theta_L)
=\frac{1}{L}\sum_{\ell=1}^L W_p^p\!\big((\theta_\ell)_\sharp \mu,\,(\theta_\ell)_\sharp \nu\big).
\label{eq:sw_estimator}
\end{equation}
For a fixed budget of $L$ slices, the quality of a SW estimate is therefore determined by the set $\Theta_L\subset\mathbb{S}^{d-1}$.
For notational convenience, we denote $f(\theta;\mu,\nu) \coloneq W_p^p(\theta_\sharp \mu,\,\theta_\sharp \nu)$.

In the absence of any information, a natural choice for selecting the $\Theta_L$ is random Monte Carlo (MC) sampling, which famously suffers from a slow $\mathcal{O}(L^{-1/2})$ error convergence rate \citep{nadjahi2020statistical}.
To mitigate this issue, recent work has applied quasi--Monte Carlo (QMC) sampling techniques to estimate SW \citep{quasiSW}.
However, QMC methods still sample pseudo-randomly, merely using data-independent heuristics to generate sample sets with more uniform coverage of the space being sampled (i.e., lower discrepancy) \citep{Niederreiter1992,dick2010}.
%of samples across the 
%(deterministic, low-discrepancy point sets) tighten this approximation when used to define Quasi--Sliced Wasserstein (QSW) estimators
Randomized counterparts (RQSW) restore unbiasedness for stochastic optimization while retaining uniformity~\citep{quasiSW}.
%Beyond Sobol/gaussianized designs, the QSW literature studies structured direction sets on $\mathbb{S}^2$ such as equal-area, spiral sequences, and energy/distance-optimized sets (e.g., Coulomb or pairwise-distance criteria), which further improve coverage and variance in practice~\citep{quasiSW}. QMC’s discrepancy guarantees explain improved absolute error, while RQSW’s randomized designs admit unbiased estimators and remain practical in 3D~\citep{quasiSW}.

However, in estimating SW, there is \textit{not} an absence of information; the motivating idea of our paper is that when computing SW, \textit{every slice yields additional information} that can inform the selection of further slices.
For instance, if two nearby $\theta$ have substantially different values of $f(\theta; \mu, \nu)$, one may wish to sample additional slices between those two slices.
Put another way, given a fixed budget of slices (especially for small $L$), the goal of selecting slices in SW should not be uniform coverage, but rather to maximize coverage where most of the ``signal'' of SW is determined.

%\paragraph{This paper.} 
Based on this insight, the present paper develops a novel family of algorithms for estimating SW, learning to select projection directions via Bayesian Optimization (BO), which proves to be especially effective in settings where SW is called inside an optimization loop (e.g., gradient-flow-style updates).
BO has become a popular, sample-efficient strategy for selecting informative queries under tight budgets across hyperparameter tuning, experimental design, and robotics~\citep{shahriari2016,snoek2012,garnett2023,uuv}.
Unlike QSW, which provides task-agnostic uniform coverage of the sphere, BO can exploit structure in the projection landscape to prioritize informative directions and adapt them as the task evolves.
Our thesis is that when the objective repeatedly queries SW on evolving distributions, a small set of task-adapted directions can accelerate convergence without altering downstream losses or gradients---and that BO provides a simple, black-box way to pick them.

Of course, BO is not the only possibility for adaptive refinement of SW estimates.
For instance, control variates such as the up/low method of \citet{nguyen2024slicedwassersteinestimationcontrol} or spherical harmonics as in \citet{leluc2024slicedwassersteinestimationsphericalharmonics} have been shown to improve MC convergence in the context of SW.
Repulsive point processes for MC have also been applied to SW \citep{petrovic2025repulsivemontecarlosphere}, showing potential advantages in higher dimensions.
Random-path projecting directions \citep{10.5555/3692070.3693608} and Markovian SW \citep{10.5555/3666122.3667852} both seek to select informative directions and show improved performance over baseline MC SW.
Nonetheless, a recent survey on sampling for sliced OT \citep{sisouk2025usersguidesamplingstrategies} confirms BO remains unexplored in this area.

%To test this capability directly, we first validate BO's adaptivity on controlled synthetic fitness landscapes, where it consistently outperforms QSW and RQSW baselines in identifying high-reward directions under tight budgets (see Section~\ref{sec:synthetic}).

Our paper ultimately introduces and evaluates four drop-in, BO-based direction selectors: \textbf{BOSW}: a one-shot BO search on $\mathbb{S}^{d-1}$ to pick $L$ directions; \textbf{RBOSW}: a periodic-refresh variant that reuses BO for light retuning during optimization; \textbf{ABOSW}: an adaptive hybrid that seeds from strong QSW sets and applies a few lightweight BO refinements and \textbf{ARBOSW}: a restarted hybrid that periodically relearns directions (fresh BO) while retaining QSW seeding.

We note that our idea of leveraging BO can be complementary to QSW approaches: for instance, combining QSW and BO yields our ABOSW and ARBOSW methods, which often achieve the best performance in our numerical results.
RQSW-style randomization can still be layered when unbiased stochastic gradients are required (under the constructions proposed in~\citet{quasiSW}).

% TODO move this to experiments section
%\paragraph{Positioning w.r.t. prior work.} The QSW study established: (i) practical QMC constructions on $\mathbb{S}^{d-1}$; (ii) deterministic QSW estimators; and (iii) RQSW estimators that are unbiased and efficient; validating these on approximation error and applications (interpolation, color transfer, autoencoders)~\citep{quasiSW}. We keep the same experimental protocol and setup and evaluate our BO-based selectors as a drop-in replacement for the direction sets used there, reporting the same metrics for fair comparison.

%\paragraph{Contributions.}
The contributions of our work include:
\begin{enumerate}
    \item We provide a family of simple, plug-and-play selectors for SW projections driven by BO.  In appropriate settings, these methods achieve state-of-the-art efficiency for estimating SW.
    \item We integrate these methods into the QSW/RQSW pipeline without changing downstream losses/gradients.% and preserve unbiasedness when paired with RQSW sampling. %~\citep{quasiSW}.
    \item On the QSW paper's test suite \citep{quasiSW}, our hybrid variants (ABOSW/ARBOSW) show convergence competitive with the best QSW baselines at modest overhead; we follow their reporting protocol for one-to-one comparability.
\end{enumerate}

\section{Background}

\paragraph{Wasserstein and sliced Wasserstein.}
Let $\mathcal{P}_p(\mathbb{R}^d)$ be the set of probability measures with finite $p$-th moment.
The $p$-Wasserstein distance between measures $\mu,\nu \in \mathcal{P}_p(\mathbb{R}^d)$ is
\[
W_p^p(\mu,\nu) \;=\; \inf_{\pi \in \Pi(\mu,\nu)} \int \|x-y\|^p \,\mathrm{d}\pi(x,y),
\]
where $\Pi(\mu,\nu)$ denotes couplings with marginals $\mu,\nu$~\citep{peyre2019computational}.
Computing $W_p$ in high dimension can be costly. The sliced Wasserstein (SW) distance mitigates this by averaging one-dimensional Wasserstein costs over directions on the unit sphere (see Equation~\ref{eq:sw-continuous}).

For empirical measures with the same number $n$ of points, each 1-D cost is the average between \emph{sorted} projections.
Let $X=\{x_i\}_{i=1}^n$ and $Y=\{y_i\}_{i=1}^n \subset \mathbb{R}^d$ be point clouds with empirical measures
\[
\hat\mu=\tfrac{1}{n}\sum_{i=1}^n \delta_{x_i},
\qquad
\hat\nu=\tfrac{1}{n}\sum_{i=1}^n \delta_{y_i},
\]
where $\delta_{z}$ denotes the \emph{Dirac unit mass} at $z$.
For the \emph{current slice direction} $\theta\in\mathbb{S}^{d-1}$, define $T_\theta(x)=\theta^\top x$ and consider the two lists of 1-D projected values
$\{\theta^\top x_i\}_{i=1}^n$ and $\{\theta^\top y_i\}_{i=1}^n$.
The corresponding \emph{projected empirical measures} are
\[
\hat\mu_\theta := (T_\theta)_\sharp \hat\mu \;=\; \tfrac{1}{n}\sum_{i=1}^n \delta_{\theta^\top x_i},
\qquad
\hat\nu_\theta := (T_\theta)_\sharp \hat\nu \;=\; \tfrac{1}{n}\sum_{i=1}^n \delta_{\theta^\top y_i}.
\]
Writing $x_{(i)}^\theta$ and $y_{(i)}^\theta$ for these projected values sorted in increasing order (the $i$-th order statistics),
\[
W_p^p(\hat{\mu}_\theta,\hat{\nu}_\theta) \;=\; \frac{1}{n} \sum_{i=1}^n \big|x_{(i)}^\theta - y_{(i)}^\theta\big|^p.
\]
Sorting is necessary because in one dimension with equal weights, the optimal transport pairs matching quantiles (monotone rearrangement); without sorting, one would either mismatch points (higher cost) or solve a full assignment problem.
Computing the 1-D cost therefore requires two sorts ($\mathcal{O}(n\log n)$) followed by a linear-time sum, yielding $\mathcal{O}(n \log n)$ time and $\mathcal{O}(n)$ memory per direction~\citep{bonneel2015,peyre2019computational}.

\paragraph{Approximating the sphere integral.}
As discussed in Section~\ref{sec:intro}, we use the finite-direction estimator in Equation~\ref{eq:sw_estimator} with a set of slices
$\Theta_L=\{\theta_\ell\}_{\ell=1}^L$; here we take $\mu=\hat\mu$ and $\nu=\hat\nu$ (empirical measures).
With i.i.d.\ Monte Carlo (MC) directions, the root mean square error scales as $\mathcal{O}(L^{-1/2})$.
Quasi--Monte Carlo (QMC) replaces random draws by low-discrepancy point sets~\citep{Niederreiter1992,dick2010}, improving integration error for sufficiently smooth integrands.

% \paragraph{Approximating the sphere integral.}
% As discussed in Section~\ref{sec:intro}, we use the finite-direction estimator in Equation~\ref{eq:sw_estimator} with a set of slices
% $\Theta_L=\{\theta_\ell\}_{\ell=1}^L$; here we take $\mu=\hat\mu$ and $\nu=\hat\nu$ (empirical measures).
% With i.i.d.\ Monte Carlo (MC) directions, the root mean square error scales as $\mathcal{O}(L^{-1/2})$.
% Quasi--Monte Carlo (QMC) replaces random draws by low-discrepancy point sets~\citep{Niederreiter1992,dick2010}, improving integration error for sufficiently smooth integrands.

\paragraph{QSW and RQSW.}
QSW instantiates QMC on $\mathbb{S}^{2}$ via (i) equal-area transforms of Sobol points, (ii) Gaussianized Sobol points projected to the sphere, and (iii) structured deterministic sets such as spiral sequences and energy/distance-optimized designs; it also studies randomized counterparts (RQSW) that either scramble the net or apply random rotations, which restore unbiasedness while retaining uniform coverage~\citep{quasiSW}. Since these algorithms have state-of-the-art performance, we evaluate our methods against the same experimental protocols (datasets, metrics, and reporting) as \citet{quasiSW} for fair comparison.

\paragraph{Bayesian optimization (BO) in a nutshell.}
BO is a state-of-the-art sample-efficient strategy for optimizing expensive, black-box functions. A surrogate (often a Gaussian process) models the objective from queried inputs; an acquisition rule (e.g., upper confidence bound (UCB) or expected improvement (EI)) trades off exploration vs.\ exploitation to select the next query.  For details, we refer to \citet{shahriari2016,snoek2012,garnett2023,uuv}.
%BO has seen broad success in hyperparameter tuning, experimental design, and robotics under tight evaluation budgets.

\paragraph{Problem view for this work.}
When SW is used inside an outer optimization (e.g., gradient flows or registration), the same direction set $\Theta_L$ is queried repeatedly on evolving point clouds.
This motivates \emph{learning} $\Theta_L$ for the task at hand.
The four method variants in our paper
%Our methods \textbf{BOSW} (one-shot), \textbf{RBOSW} (periodic refresh), \textbf{ABOSW} (QSW-seeded with light BO refinement), and \textbf{ARBOSW} (restarted QSW+BO)
operate as \emph{selectors of directions} and slot into the exact same SW pipelines as in~\citet{quasiSW}.
They do not alter the downstream objectives or gradients; and in settings requiring unbiasedness, they can be paired with the randomized estimators described by~\citet{quasiSW}.

\section{Methods}

\subsection{Bayesian optimization on the sphere}
\label{sec:bo-sphere}

We treat $f(\theta;\mu,\nu)$ from Equation~\ref{eq:sw_estimator} as a black-box function on
$\mathbb{S}^{d-1}$ and fit a Gaussian–process (GP) surrogate with covariance (kernel) $k(\cdot,\cdot)$.
We use the \emph{angular RBF} kernel, i.e., a Gaussian of the spherical geodesic distance:
\begin{equation}
k(\theta,\theta')
\;=\;
\exp\!\Big(-\tfrac{1}{2}\big(\tfrac{d_{\mathbb{S}}(\theta,\theta')}{\ell}\big)^2\Big),
\qquad
d_{\mathbb{S}}(\theta,\theta')=\arccos\langle\theta,\theta'\rangle,
\label{eq:kernel}
\end{equation}
with lengthscale $\ell>0$ (set by a median heuristic over pairwise geodesic distances among the
currently evaluated directions). This isotropic (zonal) kernel on spheres is standard
\citep{Schoenberg1942PositiveDefiniteSpheres,Borovitskiy2020MaternManifolds,rasmussen2006gpml}.

At iteration $t$, with data $\mathcal{D}_t=\{(\theta_i,f(\theta_i))\}_{i=1}^{n_t}$ (so $n_t=|\mathcal{D}_t|$),
the GP posterior $(\mu_t,\sigma_t)$ defines an acquisition $\alpha_t(\theta)$; by default we use
UCB with $\beta=0.7$ \citep{srinivas2010ucb,jones1998ego,shahriari2016}.
New directions are proposed by maximizing $\alpha_t$ over a candidate pool of size $n_c$ sampled
uniformly on $\mathbb{S}^{d-1}$ (we use $n_c=4096$ unless stated), selecting a small batch $b$
(default $b=5$). Candidates are normalized to unit norm and we suppress near-duplicates by dropping
any proposal whose cosine similarity to the current set exceeds $0.98$. These small, parallelizable
rounds keep the BO overhead modest relative to SW evaluation; per round, scoring costs
$\mathcal{O}(n_c\,n_t)$ kernel evaluations and adds only $b$ new $f$-evaluations.

We performed ablations to choose the models and parameters listed here; see Appendix~\ref{sec:ablations}.
Furthermore, we discuss the theoretical guarantees on using BO for SW in Appendix~\ref{sec:theory}.

\subsection{Four Variations of Bayesian Optimization for Sliced Wasserstein}

\paragraph{BOSW: one-shot learned directions} BOSW performs a single BO run to select $L$ directions, starting from a small random initialization and iteratively expanding $\Theta_L$ until full.  
The learned set remains fixed during downstream optimization.
In terms of computational cost, each BO round requires $\mathcal{O}(n_c n_t)$ kernel evaluations, which is modest for $L$ in the hundreds.
We note that the task-adapted directions BOSW learns do not comprise an unbiased quadrature rule for the spherical average.
We include BOSW in the approximation error plot (see Section \ref{sec:approx-error}) only for completeness and do not expect it to excel there; for unbiased estimation we rely on (R)QSW.
We omit RBOSW/ABOSW/ARBOSW from that plot since their refresh/seeding mechanisms target optimization-in-the-loop rather than a one-off integral.
%\end{itemize}

\paragraph{RBOSW: periodic refresh}

RBOSW re-runs the BOSW selection process every $R$ optimization steps using updated source and target distributions (i.e., the current pair of distributions whose distance is being measured with SW at step $t$).  
Between refreshes, the selected directions remain fixed.  
This adaptation can track evolving geometry without restarting the main optimization loop.
We note that ``refresh'' here means re-running BOSW from scratch on the current data without carrying forward the previous surrogate model.

\paragraph{ABOSW: QSW-seeded, lightweight refinement}
ABOSW is a variant that uses \textit{adaptive} BO: standard GP-based BO (UCB) whose proposals are
\emph{conditioned on the slice values already observed}.
ABOSW is adaptive to the task at initialization (QSW seed $\rightarrow$ a few BO refinements), but not time-adaptive; the set is refined once and then fixed.

ABOSW begins with a strong QSW set (e.g., spiral/Coulomb) and performs a few BO rounds to
lightly adjust it:
\begin{enumerate}
  \item \textbf{Initialize the GP} by evaluating $f(\theta)$ on the QSW seed $\Theta_L$;
        these evaluations form the initial dataset $\mathcal D_0$.
  \item \textbf{Run a small number of BO rounds} (in our experiments, we run
        $r = 2$ rounds with mini-batch size $b = 5$; thus at most $br\le 10$ directions
        change when $L = 100$, i.e., $\le 10\%$ of the set). At each round:
        sample a candidate pool of size $n_c = 4096$ uniformly on $\mathbb{S}^{d-1}$,
        score by UCB (default $\beta = 0.7$), suppress near-duplicates
        (cosine similarity $>0.98$), pick $b$ proposals, and replace the $b$ worst
        directions in $\Theta_L$.
  \item \textbf{Return the refined set} $\Theta_L$; the overhead is small since only
        $br$ new directions are evaluated (mere seconds of compute, in our experiments).
\end{enumerate}

\paragraph{ARBOSW: restarted hybrid}

ARBOSW periodically restarts the ABOSW process: at fixed intervals, it re-seeds from a QSW set and runs the short BO refinement routine on current data.  
The key distinction from RBOSW is that each restart begins from a QSW initialization rather than the previous BO state, enabling both periodic adaptation and consistent seeding.

\paragraph{Compatibility with randomized estimators}

When unbiased stochastic gradients are needed, randomized QSW remains appropriate~\citep{quasiSW}.  
Our BO-based selectors are orthogonal to this choice: QSW (deterministic or randomized) provides the base set, and BO offers refinement or re-selection.  
We do not claim unbiasedness for BOSW-based sets (cf.\ Appendix~\ref{sec:theory}); evaluations follow the same deterministic vs. randomized protocols as~\citet{quasiSW}.

% \subsection{Implementation notes}

% \begin{itemize}
%     \item \textbf{Kernel:} Angular RBF with lengthscale chosen via the median heuristic on observed angular distances.
%     \item \textbf{Acquisition:} UCB (default) or EI, with very small batch sizes and candidate pools in the low thousands.
%     \item \textbf{Safeguards:} Duplicate suppression by cosine thresholding, normalization to unit vectors.
%     \item \textbf{Complexity:} The extra cost from BO is generally a small fraction of the cost of computing SW itself.
%     \item \textbf{Evaluation protocol:} All experiments follow the datasets, metrics, and reporting format from~\citet{quasiSW} for one-to-one comparability.
% \end{itemize}

\section{Experiments}

\begin{figure}[!t]
    \centering
    \includegraphics[width=0.9\linewidth]{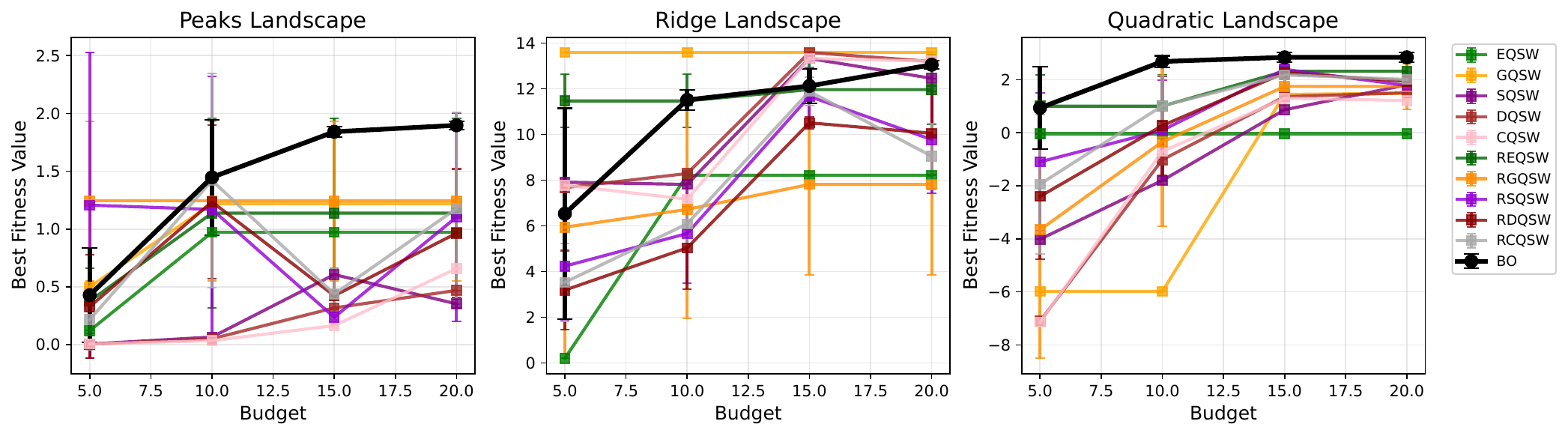}\\[-10pt]
    \caption{
    Synthetic projection-selection experiment on three fitness landscapes (higher is better). Bayesian optimization consistently and quickly finds the optimal projection directions. In the ridge example, one of the (pseudo-random) QSW methods happens to identify a useful slice early on; but BO clearly succeeds across all test cases.
    }
    \label{fig:landscapes}
\end{figure}

Aside from Section \ref{sec:synthetic}, our experiments match those run in \citet{quasiSW}: we use the same datasets, tasks, and reporting style, and we only ``swap'' the projection selector (ours vs.\ theirs).
Across all tasks, we keep the optimizer, learning rates, and stopping criteria identical to \citet{quasiSW} to ensure one-to-one comparability. We include the QSW families used in the paper, as well as vanilla Monte Carlo SW.

\subsection{Synthetic Projection-Selection Experiment}
\label{sec:synthetic}

\paragraph{Settings.}
To verify that Bayesian optimization (BO) can adaptively identify high-quality projection directions, 
we construct synthetic ``fitness landscapes'' over the sphere where certain directions yield much higher rewards. 
We design three landscapes: (i) \emph{Peaks}, with several local maxima and one dominant peak; 
(ii) \emph{Ridge}, where projections aligned with $[1,1,1]$ achieve high value; and 
(iii) \emph{Quadratic}, where a single target direction maximizes fitness. 
Each method is given a small evaluation budget $L \in \{5,10,15,20\}$ and must return the best projection found. 
We compare BO against QSW baselines (EQSW, GQSW, SQSW, DQSW, CQSW) and their randomized counterparts (RQSW, RGQSW, RSQSW, RDQSW, RCQSW) (see \citet{quasiSW} for details of each variant). 
Performance is measured by the mean best fitness value over $5$ trials. 

\paragraph{Results.}
Figure~\ref{fig:landscapes} summarizes the outcomes. 
On the \emph{Peaks} landscape, BO steadily improves with budget and clearly outperforms all QSW variants at $L=15$ and $20$ (best value $\approx 1.90$ versus $\approx 1.2$ for the strongest QSW).  
On the \emph{Ridge} landscape, QSW methods that happen to align with the ridge (notably GQSW) perform well, but BO rapidly adapts to approach the optimum, surpassing randomized variants at all budgets.  
On the \emph{Quadratic} landscape, BO dominates: it already finds near-optimal projections by $L=10$ (value $2.69$ vs.~$\leq 1.0$ for QSW/RQSW), and reaches the global optimum $\approx 2.84$ by $L=15$, while QSW remains below $2.4$.  
These results illustrate how BO leverages feedback to focus search on promising regions, whereas QSW's fixed designs (though effective for uniform coverage) cannot adapt to landscape structure. 
This synthetic experiment confirms BO's capability as an adaptive projection selector, motivating its use in sliced Wasserstein computations.

\subsection{Approximation Error}
\label{sec:approx-error}

\paragraph{Setting.}
We select four point clouds (indices 1--4; 2048 points; 3D) from the ShapeNet Core-55 \citep{chang2015} and use them as in \citet{quasiSW}.
The population (ground truth) SW value for each pair (1--2, 1--3, 2--4, 3--4) is approximated by a high-budget MC estimate with $L = 100{,}000$.
$L$ ranges from $10$ to $10,000$.  We report the absolute error of:
(i) MC (SW), 
(ii) the QSW variants (GQSW, EQSW, SQSW, DQSW, CQSW), and 
(iii) our BOSW (one-shot BO).
Stochastic methods (SW, BOSW) are averaged over five seeds; QSW variants are deterministic.

\begin{figure}[b]
    \centering
    % Top: 3D point cloud objects
    \includegraphics[width=0.75\linewidth]{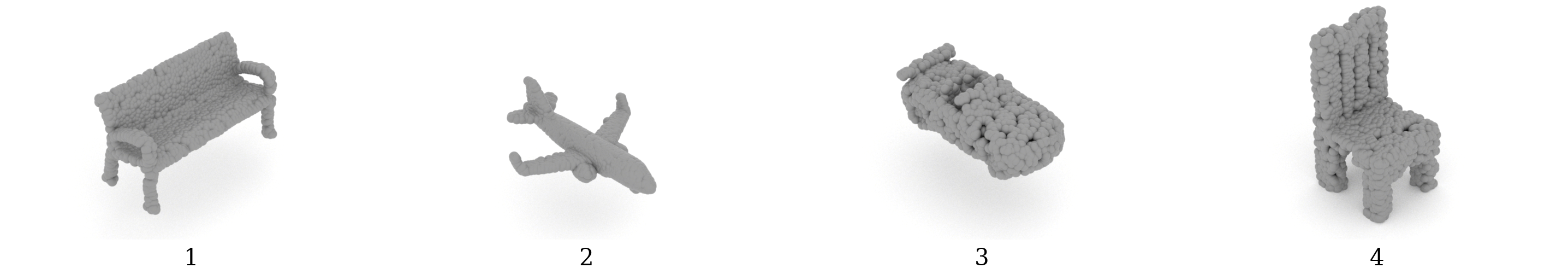}\\[3pt]
    % Bottom: Approximation error plot
    \includegraphics[width=1\linewidth]{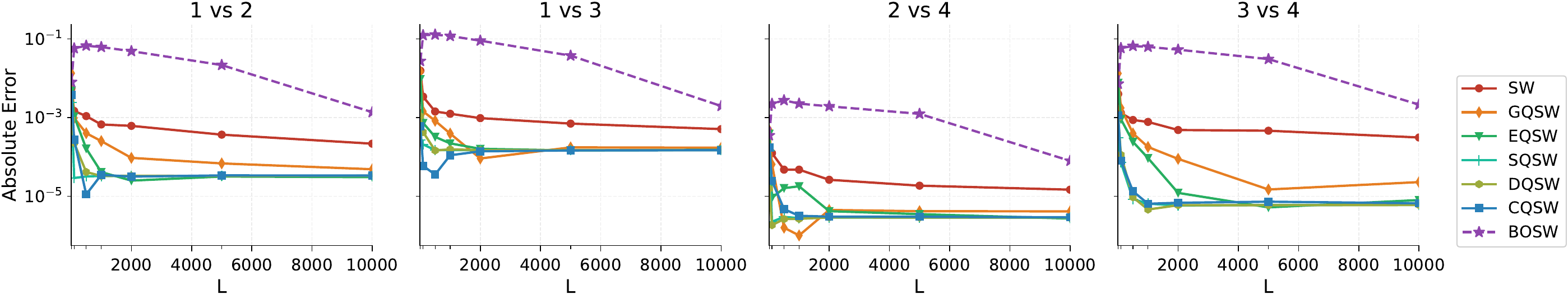} \\[-5pt]
    \caption{Approximation error for SW between empirical distributions over point clouds. BO does not provide an advantage compared to QSW methods, which excel on this type of problem that relies on uniformly sampling the sphere of possible directions.}
    \label{fig:approx_error}
\end{figure}

\paragraph{Results.}
Figure~\ref{fig:approx_error} shows absolute error versus $L$ across the four pairs.
Unsurprisingly, QSW variants consistently yield lower errors than standard MC, with CQSW/DQSW (and SQSW/EQSW closely behind) approaching the $\approx 10^{-5}$ regime once $L \gtrsim 10^3$.
In contrast, BOSW exhibits notably larger error across all $L$ and pairs, decreasing with $L$ but remaining above MC.
This is expected: in this problem, the data are highly uniform, such that slices along each direction are roughly equally meaningful.
Since BOSW learns task-adapted directions and is not designed to uniformly integrate the sphere, it does not perform competitively on approximation error for a task of this nature.
Consequently, in subsequent sections, we focus on optimization-in-the-loop settings (e.g., gradient flows), where learned directions via BO are shown to help.

\subsection{Point-cloud interpolation}
\label{sec:pci}

\paragraph{Setting.}
We evolve
\[
\dot Z(t) \;=\; -\,n\,\nabla_{Z(t)}\!\left[\mathrm{SW}_2\!\big(P_{Z(t)},\, P_Y\big)\right]
\]
to interpolate between two point clouds $X$ and $Y$.
Here, $P_X,P_Y$ are empirical distributions over $X$ and $Y$, and the curve starts at $Z(0)=X$ (and ends at $Y$).
We use the same Euler scheme as \citet{quasiSW}: $500$ iterations with step size $0.01$, and evaluate the distance between $P_{Z(t)}$ and $P_Y$ using the 2-Wasserstein metric from POT~\citep{flamary2021pot}.
We set $L=100$ directions for all methods and report the mean $\pm$ standard deviation over three seeds at steps $t\in\{100,200,300,400,500\}$.
We compare SW, the QSW family, the randomized QSW (RQSW) family, and our BO-driven selectors: BOSW (one-shot), RBOSW (periodic refresh), ABOSW (hybrid seed+refine), and ARBOSW (restarted hybrid).
%Table~\ref{tab:gf_w2} shows the performance for specific variants.

\paragraph{Results.}
Table~\ref{tab:gf_w2} reports $W_2$ ($\times 10^2$) and wall-clock time. As in \citet{quasiSW}, randomized QSW variants generally yield the shortest trajectories at later steps.
However, our \textbf{ARBOSW} achieves the \emph{best} value at step 500 (same mean as REQSW/RCQSW but smaller variance), with a moderate overhead (7.1\,s vs 4.0–4.3\,s for QSW baselines).
\textbf{RBOSW} is strongest early (best at steps 100 and 200) but incurs large cost due to periodic BO refreshes.
The deterministic one-shot \textbf{BOSW} trails (consistent with the need for fresh directions as the flow evolves), while \textbf{ABOSW} performs similarly to the deterministic QSW group.
Overall, BO hybrids provide competitive convergence without changing the optimization loop.
For completeness, we also report results at a much smaller budget ($L=10$ directions) in Appendix~\ref{app:gf_l10}. The trends are consistent: ARBOSW remains among the top-performing estimators, RBOSW dominates early steps but is slower, and BOSW/ABOSW align with their deterministic counterparts.

\begin{table}[t]
\centering
\small
\resizebox{0.85\linewidth}{!}{%
\setlength{\tabcolsep}{6pt}
\begin{tabular}{lccccc c}
\toprule
\textbf{Estimators} & \textbf{Step 100 (W$_2 \downarrow$)} & \textbf{Step 200 (W$_2 \downarrow$)} & \textbf{Step 300 (W$_2 \downarrow$)} & \textbf{Step 400 (W$_2 \downarrow$)} & \textbf{Step 500 (W$_2 \downarrow$)} & \textbf{Time (s$\downarrow$)} \\
\midrule
SW              & 5.641 $\pm$ 0.092 & 0.162 $\pm$ 0.002 & 0.025 $\pm$ 0.002 & 0.010 $\pm$ 0.002 & 0.005 $\pm$ 0.001 & 3.81 \\
\midrule
GQSW            & 6.012 $\pm$ 0.000 & 0.242 $\pm$ 0.000 & 0.074 $\pm$ 0.000 & 0.067 $\pm$ 0.000 & 0.066 $\pm$ 0.000 & 3.85 \\
EQSW            & 5.316 $\pm$ 0.000 & 0.208 $\pm$ 0.000 & 0.068 $\pm$ 0.000 & 0.062 $\pm$ 0.000 & 0.062 $\pm$ 0.000 & 4.07 \\
SQSW            & 5.622 $\pm$ 0.000 & 0.186 $\pm$ 0.000 & 0.085 $\pm$ 0.000 & 0.077 $\pm$ 0.000 & 0.075 $\pm$ 0.000 & 4.14 \\
DQSW            & 5.681 $\pm$ 0.000 & 0.184 $\pm$ 0.000 & 0.079 $\pm$ 0.000 & 0.072 $\pm$ 0.000 & 0.069 $\pm$ 0.000 & 4.26 \\
CQSW            & 5.515 $\pm$ 0.000 & 0.169 $\pm$ 0.000 & 0.077 $\pm$ 0.000 & 0.071 $\pm$ 0.000 & 0.070 $\pm$ 0.000 & 4.24 \\
\midrule
RGQSW           & 5.620 $\pm$ 0.022 & 0.161 $\pm$ 0.001 & \textbf{0.022} $\pm$ 0.003 & 0.009 $\pm$ 0.002 & 0.004 $\pm$ 0.001 & 4.04 \\
RRGQSW          & 5.628 $\pm$ 0.024 & 0.159 $\pm$ 0.005 & 0.025 $\pm$ 0.003 & 0.010 $\pm$ 0.002 & 0.004 $\pm$ 0.001 & 4.01 \\
REQSW           & 5.615 $\pm$ 0.022 & 0.160 $\pm$ 0.000 & 0.023 $\pm$ 0.002 & 0.008 $\pm$ 0.002 & 0.003 $\pm$ 0.001 & 4.15 \\
RREQSW          & 5.633 $\pm$ 0.021 & 0.159 $\pm$ 0.003 & 0.023 $\pm$ 0.001 & 0.010 $\pm$ 0.001 & 0.005 $\pm$ 0.000 & 4.12 \\
RSQSW           & 5.623 $\pm$ 0.001 & 0.162 $\pm$ 0.003 & 0.024 $\pm$ 0.003 & 0.009 $\pm$ 0.001 & 0.004 $\pm$ 0.000 & 4.17 \\
RDQSW           & 5.623 $\pm$ 0.005 & 0.163 $\pm$ 0.003 & 0.025 $\pm$ 0.003 & 0.009 $\pm$ 0.001 & 0.004 $\pm$ 0.000 & 4.28 \\
RCQSW           & 5.621 $\pm$ 0.006 & 0.161 $\pm$ 0.004 & 0.023 $\pm$ 0.001 & \textbf{0.008} $\pm$ 0.001 & 0.003 $\pm$ 0.001 & 4.20 \\
\midrule
\textbf{BOSW (ours)}     & 6.096 $\pm$ 1.270 & 1.088 $\pm$ 0.495 & 0.337 $\pm$ 0.118 & 0.231 $\pm$ 0.057 & 0.196 $\pm$ 0.039 & 3.99 \\
\textbf{RBOSW (ours)} & \textbf{4.155} $\pm$ 0.226 & \textbf{0.115} $\pm$ 0.001 & 0.034 $\pm$ 0.003 & 0.024 $\pm$ 0.003 & 0.018 $\pm$ 0.002 & 44.45 \\
\textbf{ABOSW (ours)}    & 6.532 $\pm$ 0.662 & 0.315 $\pm$ 0.082 & 0.089 $\pm$ 0.007 & 0.078 $\pm$ 0.002 & 0.076 $\pm$ 0.002 & 3.86 \\
\textbf{ARBOSW (ours)} & 5.690 $\pm$ 0.081 & 0.167 $\pm$ 0.005 & 0.025 $\pm$ 0.001 & 0.009 $\pm$ 0.000 & \textbf{0.003} $\pm$ 0.000 & 7.11 \\
\bottomrule
\end{tabular}%
} % end resizebox
\caption{Summary of Wasserstein-2 distances (multiplied by $10^2$) from three different runs (L=100).}
\label{tab:gf_w2}
\end{table}

\subsection{Image Style Transfer}
\label{sec:ist}

\paragraph{Setting.}
RGB source and target image are treated as point clouds
$X,Y\in\mathbb{R}^{n\times 3}$ ($n$ is the number of pixels) and are evolved along the same SW–driven curve as in Section~\ref{sec:pci}.
%$\dot Z(t)=-\,n\,\nabla_{Z(t)}\!\big[\mathrm{SW}_2(P_{Z(t)},P_Y)\big]$,
%starting at $Z(0)=X$ and ending at $Y$.
As in \citet{quasiSW}, we round RGB values to $\{0,\ldots,255\}$ at the final Euler step, use $1{,}000$ iterations with step size $1$, and set $L=100$ projection directions for all methods; $W_2$ is computed with POT~\citep{flamary2021pot}.
We visualize SW, \textbf{RCQSW}, and our \textbf{RBOSW}/\textbf{ARBOSW}.
We report RCQSW as the representative RQSW baseline because the QSW study found it consistently among the top performers.
Randomized variants behave very similarly overall; see Appendix~\ref{app:autoencoder_full} for full results.

\begin{figure}[H]
  \centering
  \includegraphics[width=0.9\linewidth]{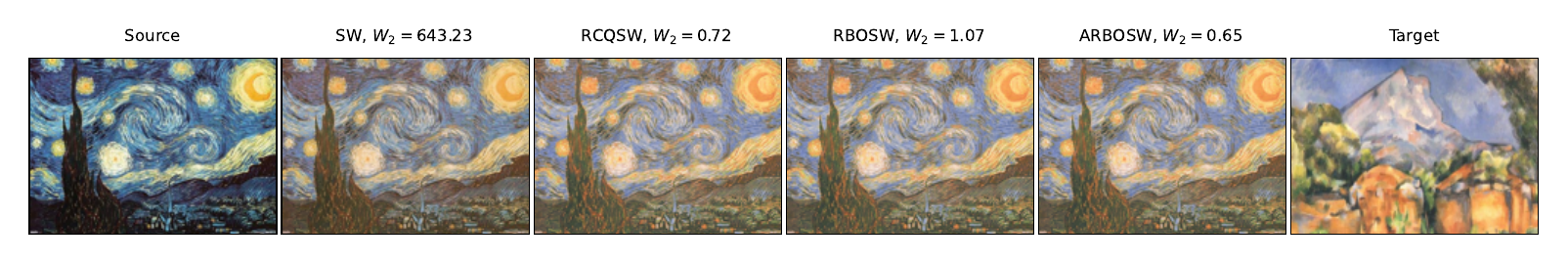} \\[-13pt]
  \caption{Image style transfer with $L{=}100$ and $1{,}000$ iterations.}
  \label{fig:style_transfer}
\end{figure}

\paragraph{Results.}
Figure~\ref{fig:style_transfer} shows a representative transfer.
Both RCQSW and our BO hybrids clearly outperform vanilla SW in terms of final $W_2$ (printed above each panel) and visual fidelity.
For ease of exposition, we keep \textbf{RCQSW} as the representative randomized baseline in the main text, mirroring the recommendation of \citet{quasiSW}; see Appendix~\ref{app:style_transfer} for full comparisons.
%and yields clean, readable figures.
We note, for transparency, that other randomized variants can be marginally stronger on some instances; in particular, \textbf{RGQSW} achieves the lowest final $W_2$ on our example at both $L{=}10$ and $L{=}100$.
However, the randomized family behaves very similarly overall, and RCQSW remains competitive.
Our \textbf{ARBOSW} matches RCQSW on this example while preserving contrast and texture; \textbf{RBOSW} improves over SW but typically trails RCQSW/ARBOSW, consistent with its lighter periodic refresh.
Thus, while not as clear of an accuracy advantage as seen in Sections \ref{sec:pci} and \ref{sec:dpca}, ARBOSW is at least competitive with the state-of-the-art for this example.

\subsection{Deep Point-cloud Autoencoder}
\label{sec:dpca}

\paragraph{Setting.}
Following \citet{nguyen2023} and \citet{quasiSW}, we train deep point-cloud autoencoders with SW on ShapeNet Core-55~\citep{chang2015}.
This amounts to optimizing an objective function:
\[
\min_{\phi, \psi} \ \mathbb{E}_{X \sim \mu(X)} \big[ \mathrm{SW}_p\!\left( P_X,\, P_{g_\psi(f_\phi(X))} \right) \big],
\]
where $\mu(X)$ is the data distribution, $f_\phi$ is a deep encoder, and $g_\psi$ is a deep decoder.  Both $f_\phi$ and $g_\psi$ use a PointNet architecture~\citep{qi2017pointnet}.
We approximate gradients using either MC, QSW, RQSW, or our BO-driven selectors, and train for $400$ epochs with SGD (learning rate $10^{-3}$, batch size $128$, momentum $0.9$, weight decay $5\times 10^{-4}$).
We set $L=100$ directions for all methods.
Evaluation is on the distinct ModelNet40 dataset~\citep{wu2015modelnet}; we report the mean reconstruction loss over three runs, measured against $W_2$ and $\mathrm{SW}_2$ (which are estimated via $10{,}000$ MC projections).

\begin{wrapfigure}{r}{0.6\textwidth}
  \centering
  \includegraphics[width=0.6\textwidth]{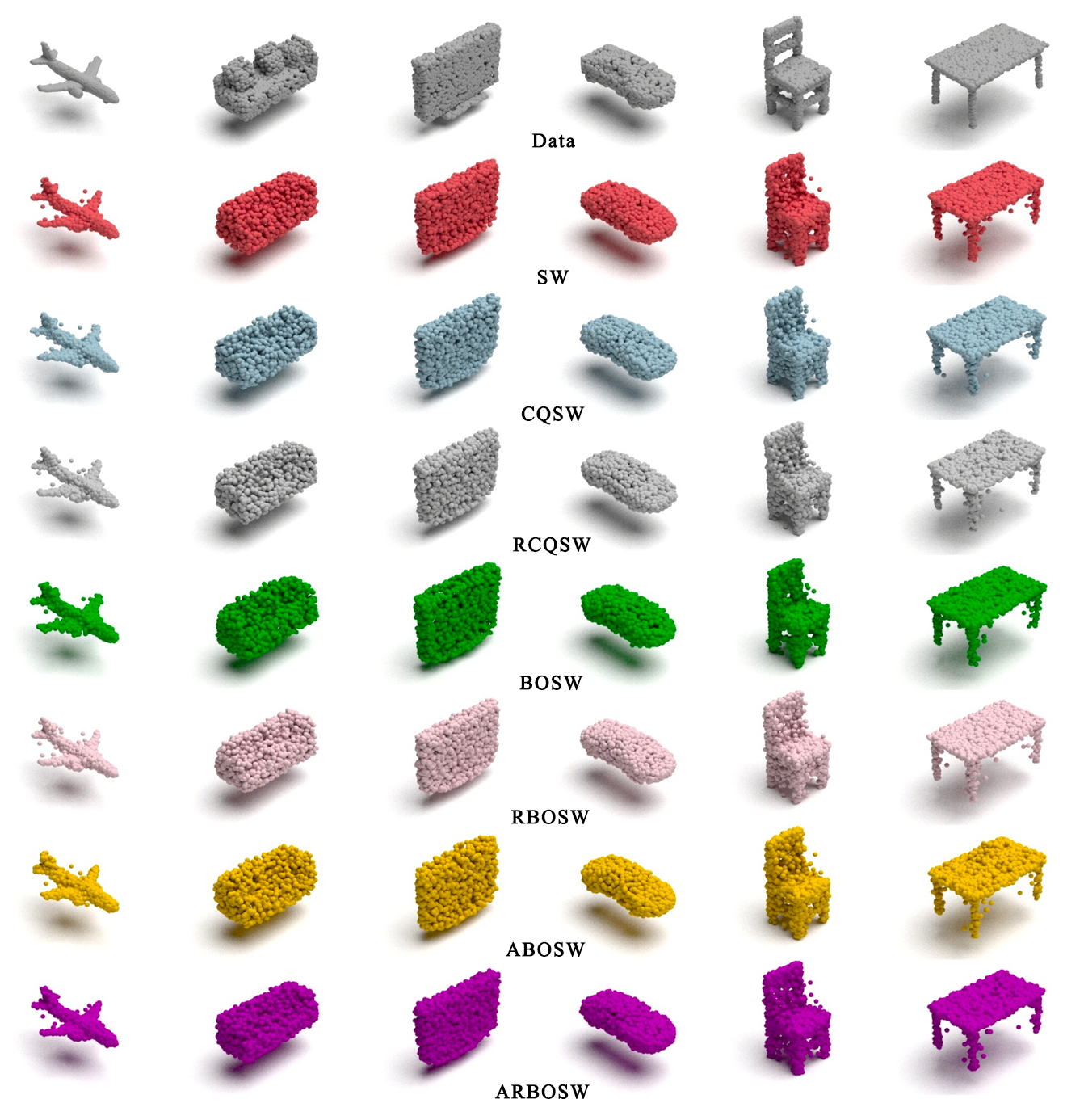} \\[-10pt]
  \caption{Reconstructed point clouds from the deep autoencoder experiment, with $L = 100$.}
  \label{fig:ae_recon}
\end{wrapfigure}

\paragraph{Results.}
Table~\ref{tab:autoencoder} and Figure~\ref{fig:ae_recon} show the reconstruction losses and qualitative outputs with $L=100$.
Consistent with prior work, CQSW is a strong deterministic baseline. However, our BO-based methods, particularly \textbf{ABOSW}, achieve the lowest reconstruction losses across epochs, even surpassing CQSW in both $\mathrm{SW}_2$ and $W_2$ at the final epoch.
BOSW also performs competitively, while the restart variants (RBOSW, ARBOSW) lag slightly behind their non-restart counterparts.
This reversal compared to gradient flows highlights the different nature of dataset-level training: here, the distribution is large and stable, so maintaining consistent, high-quality projection sets (as in BOSW/ABOSW) is more effective than frequent refreshes, which can inject unnecessary variability.
We emphasize that while methods like RGQSW, RRGQSW, and REQSW were top performers for image style transfer (Section \ref{sec:ist}), they are among the worst-performing methods in Table~\ref{tab:autoencoder} after 400 epochs.
Overall, BO-based selectors not only remain competitive but, in this setting, \textbf{outperform the previously recommended CQSW}, suggesting that task-adaptivity via BO can provide tangible benefits beyond the low-discrepancy sampling enabled by QSW.
For clarity, we include SW, CQSW, and our BO variants in the main text; the full visual and quantitative comparisons with all QSW/RQSW variants are deferred to Appendix~\ref{app:autoencoder_full}.

\begin{table}[H]
\centering
\small
\resizebox{0.9\linewidth}{!}{%
\setlength{\tabcolsep}{6pt}
\begin{tabular}{lcccccc}
\toprule
\textbf{Approximation} & \multicolumn{2}{c}{\textbf{Epoch 100}} & \multicolumn{2}{c}{\textbf{Epoch 200}} & \multicolumn{2}{c}{\textbf{Epoch 400}} \\
\cmidrule(r){2-3} \cmidrule(r){4-5} \cmidrule(r){6-7}
 & SW$_2$($\downarrow$) & W$_2$($\downarrow$) & SW$_2$($\downarrow$) & W$_2$($\downarrow$) & SW$_2$($\downarrow$) & W$_2$($\downarrow$) \\
\midrule
SW              & 2.25 $\pm$ 0.06 & 10.58 $\pm$ 0.12 & 2.11 $\pm$ 0.04 & 9.92 $\pm$ 0.08 & 1.94 $\pm$ 0.06 & 9.21 $\pm$ 0.06 \\
\midrule
GQSW            & 11.17 $\pm$ 0.07 & 32.58 $\pm$ 0.06 & 11.75 $\pm$ 0.07 & 33.27 $\pm$ 0.09 & 14.82 $\pm$ 0.02 & 37.99 $\pm$ 0.05 \\
EQSW            & 2.25 $\pm$ 0.02 & 10.57 $\pm$ 0.02 & 2.05 $\pm$ 0.02 & 9.84 $\pm$ 0.07 & 1.90 $\pm$ 0.04 & 9.20 $\pm$ 0.07 \\
SQSW            & 2.25 $\pm$ 0.01 & 10.57 $\pm$ 0.03 & 2.08 $\pm$ 0.01 & 9.90 $\pm$ 0.04 & 1.90 $\pm$ 0.02 & 9.17 $\pm$ 0.05 \\
DQSW            & 2.24 $\pm$ 0.07 & 10.58 $\pm$ 0.05 & 2.06 $\pm$ 0.04 & 9.83 $\pm$ 0.01 & 1.86 $\pm$ 0.05 & 9.12 $\pm$ 0.07 \\
CQSW            & 2.22 $\pm$ 0.02 & 10.54 $\pm$ 0.02 & 2.05 $\pm$ 0.06 & 9.81 $\pm$ 0.04 & 1.84 $\pm$ 0.02 & 9.06 $\pm$ 0.02 \\
\midrule
RGQSW           & 2.25 $\pm$ 0.02 & 10.57 $\pm$ 0.01 & 2.09 $\pm$ 0.03 & 9.92 $\pm$ 0.01 & 1.94 $\pm$ 0.02 & 9.18 $\pm$ 0.02 \\
RRGQSW          & 2.23 $\pm$ 0.01 & 10.51 $\pm$ 0.04 & 2.06 $\pm$ 0.05 & 9.84 $\pm$ 0.06 & 1.88 $\pm$ 0.09 & 9.16 $\pm$ 0.11 \\
REQSW           & 2.24 $\pm$ 0.04 & 10.53 $\pm$ 0.04 & 2.08 $\pm$ 0.04 & 9.90 $\pm$ 0.08 & 1.89 $\pm$ 0.04 & 9.17 $\pm$ 0.06 \\
RREQSW          & 2.21 $\pm$ 0.04 & 10.50 $\pm$ 0.04 & 2.03 $\pm$ 0.02 & 9.83 $\pm$ 0.02 & 1.88 $\pm$ 0.05 & 9.15 $\pm$ 0.06 \\
RSQSW           & 2.22 $\pm$ 0.05 & 10.53 $\pm$ 0.01 & 2.04 $\pm$ 0.06 & 9.82 $\pm$ 0.06 & 1.85 $\pm$ 0.05 & 9.12 $\pm$ 0.02 \\
RDQSW           & 2.21 $\pm$ 0.03 & 10.50 $\pm$ 0.02 & 2.03 $\pm$ 0.04 & 9.82 $\pm$ 0.04 & 1.86 $\pm$ 0.03 & 9.12 $\pm$ 0.02 \\
RCQSW           & 2.22 $\pm$ 0.03 & 10.50 $\pm$ 0.05 & 2.03 $\pm$ 0.02 & 9.82 $\pm$ 0.03 & 1.85 $\pm$ 0.06 & 9.12 $\pm$ 0.03 \\
\midrule
BOSW          & 2.20 $\pm$ 0.01 & 10.34 $\pm$ 0.02 & 2.02 $\pm$ 0.04 & 9.78 $\pm$ 0.03 & \textbf{1.80 $\pm$ 0.01} & \textbf{9.01 $\pm$ 0.02} \\
RBOSW          & 2.28 $\pm$ 0.03 & 10.54 $\pm$ 0.04 & 2.09 $\pm$ 0.05 & 9.84 $\pm$ 0.05 & 1.90 $\pm$ 0.02 & 9.10 $\pm$ 0.03 \\
ABOSW          & \textbf{2.18 $\pm$ 0.01} & \textbf{10.27 $\pm$ 0.02} & \textbf{2.01 $\pm$ 0.03} & \textbf{9.76 $\pm$ 0.02} & \textbf{1.81 $\pm$ 0.02} & \textbf{9.01 $\pm$ 0.03} \\
ARBOSW          & 2.21 $\pm$ 0.01 & 10.44 $\pm$ 0.02 & 2.04 $\pm$ 0.04 & 9.80 $\pm$ 0.03 & 1.85 $\pm$ 0.02 & 9.07 $\pm$ 0.02 \\
\bottomrule
\end{tabular}%
}
\caption{Reconstruction losses on the autoencoder task from different approximations with $L = 100$.  Losses are scaled by $10^2$ for ease of exposition.}
\label{tab:autoencoder}
\end{table}

% \begin{figure}[H]
%   \centering
%   \includegraphics[width=0.6\linewidth]{figs/main_ae.png} \\[-10pt]
%   \caption{Reconstructed point clouds from the deep autoencoder experiment, with $L = 100$.}
%   \label{fig:ae_recon}
% \end{figure}

\section{Concluding Remarks}

We proposed BOSW, RBOSW, ABOSW, and ARBOSW: Bayesian optimization–driven selectors for projection directions in sliced Wasserstein (SW) computations that complement quasi–Monte Carlo (QSW) designs.
Our hybrids combine QSW's low-discrepancy coverage with BO's task adaptivity, drop in to existing SW-based optimization loops without changing losses or gradients, and preserve unbiasedness when paired with randomized QSW.
Across the QSW benchmark suite of \citet{quasiSW}, ABOSW and ARBOSW match the best QSW baselines in convergence, RBOSW offers strong early-stage gains, and BOSW provides a simple deterministic alternative.
Collectively, our methods perform \textbf{competitively} on the image style transfer task, and achieve \textbf{state-of-the-art} convergence for the point-cloud interpolation and deep point-cloud autoencoder tasks.
By uniting deterministic geometric constructions with data-driven adaptivity, our work broadens the design space for SW direction sets and opens a path toward principled, learned projections in large-scale optimal transport applications.

A common criticism of BO is its poor scaling with respect to the dimension $d$ of the data.
However, recent literature has suggested several variants of BO that perform well in high dimensions \citep{kim2021deep,shen2021computationally,binois2022survey,jaquier2020high}, and in an exciting development, \citet{hvarfner2024vanillabayesianoptimizationperforms} just found that even vanilla BO can perform excellently in high dimensions with a simple scaling of the GP lengthscale.
Thus, it is no longer fair to claim that BO is cursed by dimensionality.
We would like to test our methods on higher-dimensional datasets in future work.

We remark that the present work is reminiscent of (adaptive) Bayesian quadrature (BQ) \citep{bq,10.5555/3454287.3454847,briol2015frank,activeBQ}.
However, BQ is designed to compute an integral itself, not select individual directions.
In our setting, the integrand changes every step, so BQ would require refitting a GP and recomputing quadrature weights each time, for a total of $O(L^3)$ runtime.
Thus, while BQ is infeasible to directly apply, it remains closely related to the ideas in our work.

Finally, we imagine several opportunities to improve our methods.
Parallelism and GPU implementation, as explored by others in the field \citep{balandat2020botorch,knudde2017gpflowopt,munawar2009theoretical}, could significantly accelerate the runtime of our methods.
Adding constraints or information to BO based on knowledge of the problem, as in \citet{eriksson2021scalable,gelbart2015constrained,vardhan2023constrained,jaquier2022geometry}, would give our methods further advantage over data-blind techniques like QSW.
Finally, recent variants of BO could offer higher-performance drop-in replacements for the BO functions in our paper \citep{visser2025labcat,mcleod2018optimization,kawaguchi2015bayesian}.
We aim to explore these extensions in future work.

 \subsubsection*{Acknowledgments}
 This paper was inspired by a Ph.D.\ preliminary exam of a student advised by Soheil Kolouri, where the authors found that sliced Wasserstein is about approximating integrals accurately; and by the DARPA Design.R project, which introduced the authors to Bayesian optimization.
 D.H. acknowledges Peter Volgyesi for involving him in the Design.R efforts.
 We thank the SyBBURE Searle Undergraduate Research Program for providing funding and support throughout this research.
 This material is based upon work supported by the National Science Foundation under Grant No. 2442853.

%\clearpage

%\nocite{*}

\bibliography{iclr2026_conference}
\bibliographystyle{iclr2026_conference}
\nocite{*}

\newpage
\appendix
\section*{Appendix}

\section{Additional Experimental Details and Results}

\subsection{Point-cloud interpolation with \texorpdfstring{$L=10$}{L = 10} directions}
\label{app:gf_l10}

\begin{table}[H]
\centering
\small
\resizebox{0.85\linewidth}{!}{%
\setlength{\tabcolsep}{6pt}
\begin{tabular}{lccccc c}
\toprule
\textbf{Estimators} & \textbf{Step 100 (W$_2 \downarrow$)} & \textbf{Step 200 (W$_2 \downarrow$)} & \textbf{Step 300 (W$_2 \downarrow$)} & \textbf{Step 400 (W$_2 \downarrow$)} & \textbf{Step 500 (W$_2 \downarrow$)} & \textbf{Time (s$\downarrow$)} \\
\midrule
SW              & 5.719 $\pm$ 0.113 & 0.190 $\pm$ 0.016 & 0.041 $\pm$ 0.001 & 0.020 $\pm$ 0.002 & 0.011 $\pm$ 0.001 & \textbf{2.71} \\
\midrule
GQSW            & 9.207 $\pm$ 0.000 & 3.692 $\pm$ 0.000 & 2.461 $\pm$ 0.000 & 2.191 $\pm$ 0.000 & 2.119 $\pm$ 0.000 & 2.92 \\
EQSW            & 4.045 $\pm$ 0.000 & 0.552 $\pm$ 0.000 & 0.490 $\pm$ 0.000 & 0.487 $\pm$ 0.000 & 0.482 $\pm$ 0.000 & 2.78 \\
SQSW            & 6.323 $\pm$ 0.000 & 1.044 $\pm$ 0.000 & 0.582 $\pm$ 0.000 & 0.538 $\pm$ 0.000 & 0.534 $\pm$ 0.000 & 2.84 \\
DQSW            & 5.897 $\pm$ 0.000 & 0.856 $\pm$ 0.000 & 0.612 $\pm$ 0.000 & 0.595 $\pm$ 0.000 & 0.594 $\pm$ 0.000 & 2.79 \\
CQSW            & 5.574 $\pm$ 0.000 & 0.755 $\pm$ 0.000 & 0.592 $\pm$ 0.000 & 0.582 $\pm$ 0.000 & 0.582 $\pm$ 0.000 & 2.87 \\
\midrule
RGQSW           & 6.036 $\pm$ 0.231 & 0.183 $\pm$ 0.022 & 0.034 $\pm$ 0.003 & 0.017 $\pm$ 0.003 & 0.009 $\pm$ 0.002 & 2.84 \\
RRGQSW          & 6.002 $\pm$ 0.149 & 0.215 $\pm$ 0.025 & 0.047 $\pm$ 0.004 & 0.029 $\pm$ 0.003 & 0.024 $\pm$ 0.002 & 2.78 \\
REQSW           & 5.722 $\pm$ 0.153 & 0.189 $\pm$ 0.008 & 0.039 $\pm$ 0.005 & 0.019 $\pm$ 0.004 & 0.009 $\pm$ 0.002 & 2.80 \\
RREQSW          & 5.832 $\pm$ 0.018 & 0.193 $\pm$ 0.006 & 0.038 $\pm$ 0.002 & 0.022 $\pm$ 0.001 & 0.015 $\pm$ 0.000 & 2.83 \\
RSQSW           & 5.718 $\pm$ 0.064 & 0.187 $\pm$ 0.008 & \textbf{0.035 $\pm$ 0.004} & \textbf{0.015 $\pm$ 0.002} & \textbf{0.007 $\pm$ 0.001} & 2.81 \\
RDQSW           & 5.694 $\pm$ 0.044 & \textbf{0.182 $\pm$ 0.020} & 0.033 $\pm$ 0.002 & 0.015 $\pm$ 0.001 & 0.007 $\pm$ 0.002 & 2.79 \\
RCQSW           & \textbf{5.673 $\pm$ 0.023} & 0.184 $\pm$ 0.008 & 0.038 $\pm$ 0.005 & 0.018 $\pm$ 0.003 & 0.008 $\pm$ 0.002 & 2.79 \\
\midrule
\textbf{BOSW (ours)}   & 4.833 $\pm$ 0.863 & 1.303 $\pm$ 0.494 & 0.967 $\pm$ 0.375 & 0.912 $\pm$ 0.345 & 0.900 $\pm$ 0.333 & 2.86 \\
\textbf{RBOSW (ours)}  & \textbf{3.457 $\pm$ 0.109} & 0.175 $\pm$ 0.044 & 0.082 $\pm$ 0.009 & 0.059 $\pm$ 0.005 & 0.047 $\pm$ 0.004 & 22.63 \\
\textbf{ABOSW (ours)}  & 9.086 $\pm$ 3.294 & 3.005 $\pm$ 1.499 & 1.637 $\pm$ 0.520 & 1.297 $\pm$ 0.205 & 1.216 $\pm$ 0.139 & 3.09 \\
\textbf{ARBOSW (ours)} & 5.617 $\pm$ 0.128 & 0.207 $\pm$ 0.015 & 0.039 $\pm$ 0.002 & 0.017 $\pm$ 0.003 & 0.009 $\pm$ 0.001 & 3.09 \\
\bottomrule
\end{tabular}%
} % end resizebox
\caption{Summary of Wasserstein-2 distances (multiplied by $10^2$) from three different runs (L=10).}
\label{tab:gf_w2_l10}
\end{table}

Table~\ref{tab:gf_w2_l10} reports results for point-cloud interpolation with a smaller projection budget of $L=10$. As expected, variance across methods is larger under such limited directions, but the relative trends mirror those observed for $L=100$: randomized QSW variants achieve fast convergence at later steps, while \textbf{ARBOSW} remains within the top four methods overall, offering competitive final accuracy. \textbf{RBOSW} continues to excel early in the trajectory but incurs a higher runtime due to periodic refreshes.

\subsection{Complete image style transfer comparisons}
\label{app:style_transfer}

\begin{figure}[t]
  \centering
  \includegraphics[width=\linewidth]{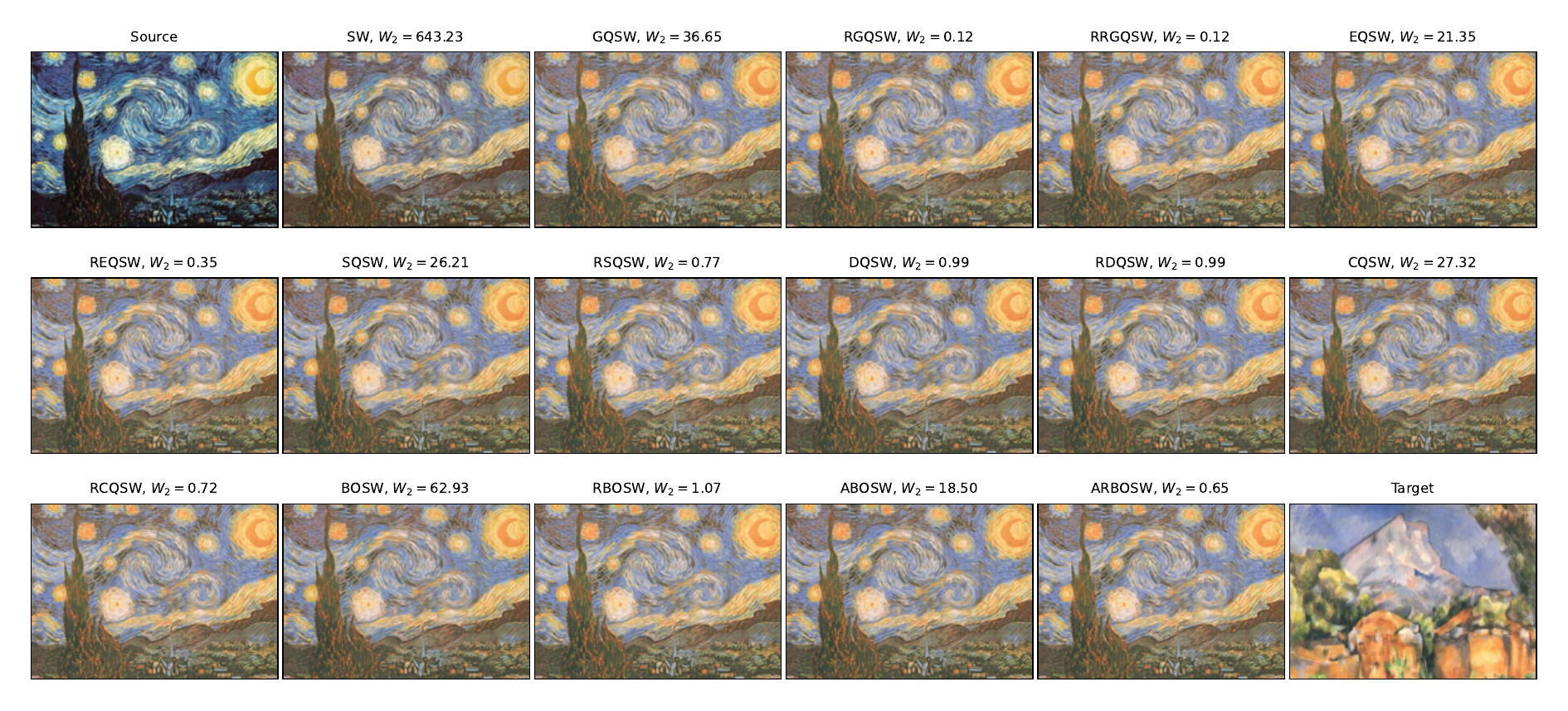}
  \caption{Full comparison for image style transfer with $L{=}100$.
  }
  \label{fig:style_transfer_all_L100}
\end{figure}

\begin{figure}[t]
  \centering
  \includegraphics[width=\linewidth]{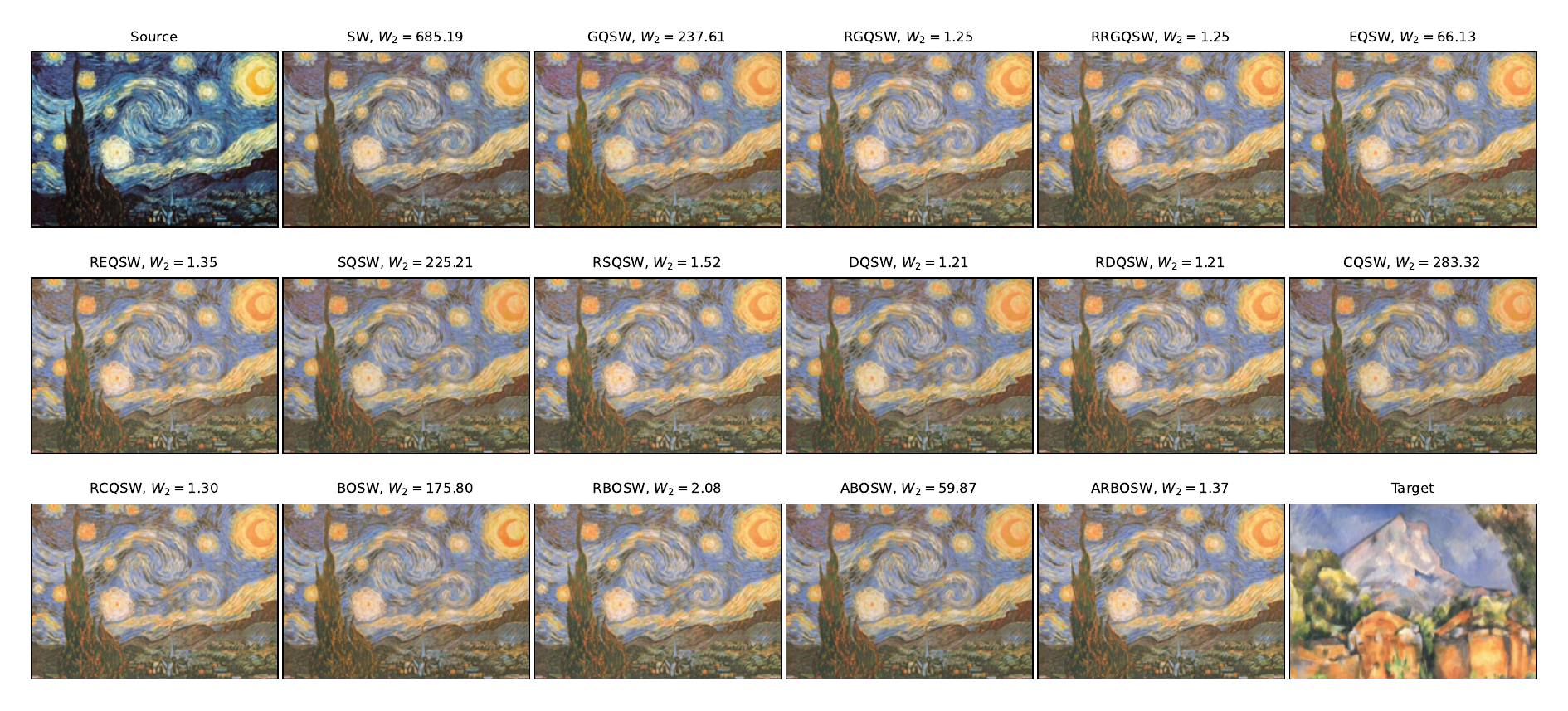}
  \caption{Full comparison for image style transfer with $L{=}10$.
  }
  \label{fig:style_transfer_all_L10}
\end{figure}

Figures~\ref{fig:style_transfer_all_L100} and~\ref{fig:style_transfer_all_L10} visualize the full method grid.
Across both budgets, the randomized family performs very similarly; within that family \textbf{RGQSW} attains the \emph{lowest} final $W_2$ on our example for \emph{both} $L{=}10$ and $L{=}100$.
This is consistent with the QSW observation that randomized designs are closely clustered in quality.
Our BO hybrids remain competitive: \textbf{ARBOSW} is comparable to RCQSW visually and in $W_2$ at $L{=}100$, while \textbf{RBOSW} improves over SW but generally trails the strongest RQSW variants, reflecting its lighter refresh.
These results support our choice to use RCQSW as the representative randomized baseline in the main figures while providing full transparency here.

\begin{figure}[hbtp]
  \centering
  \includegraphics[width=0.65\linewidth]{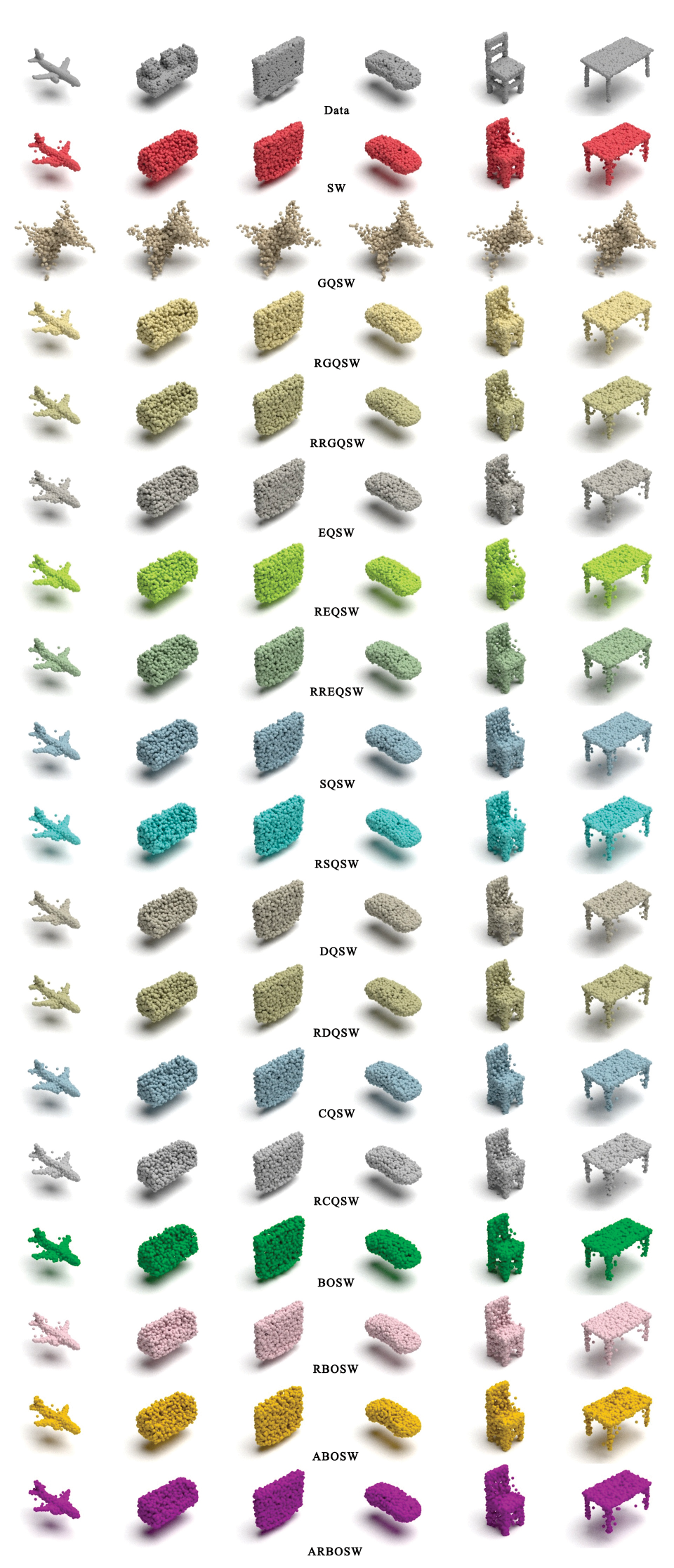} % replace with actual filename if needed
  \caption{Full reconstructed point-clouds from SW, QSW, RQSW, and BO variants with L = 100.
  }
  \label{fig:ae_appendix_full}
\end{figure}

\subsection{Full comparison for deep point-cloud autoencoders}
\label{app:autoencoder_full}

\paragraph{Setting.}
We follow the same setup as in Section~\ref{sec:dpca}: ShapeNet Core-55 training, PointNet encoder/decoder, $L=100$ projection directions, SGD with learning rate $10^{-3}$, momentum $0.9$, weight decay $5 \times 10^{-4}$, batch size $128$, and training for 400 epochs.
Evaluation is performed on ModelNet40, reporting $\mathrm{SW}_2$ and $W_2$ reconstruction losses averaged over three runs.
Here, we expand the comparison to include the entire QSW family (EQSW, GQSW, SQSW, DQSW, CQSW) and their randomized counterparts (RQSW, RRGQSW, REQSW, RREQSW, RSQSW, RDQSW, RCQSW), alongside our BO-based methods (BOSW, RBOSW, ABOSW, ARBOSW).

\paragraph{Results.}
Figure~\ref{fig:ae_appendix_full} shows qualitative reconstructions across all baselines and our BO variants.
While randomized QSW methods produce consistent and visually plausible outputs, our BO-based designs, especially \textbf{ABOSW}, stand out with sharper reconstructions and more stable geometry.
Notably, CQSW remains a strong deterministic baseline, but \textbf{ABOSW surpasses it} in final reconstruction loss while maintaining visual quality.
This contrasts with gradient flow experiments, where refresh-style variants (RBOSW/ARBOSW) had stronger performance; in autoencoders, the stable large-scale dataset favors consistent one-shot or hybrid projection sets (BOSW/ABOSW).
These observations reinforce our conclusion that the optimal projection strategy is task-dependent: stability-driven tasks benefit from non-refresh BO, while dynamic flows gain from refresh hybrids.

\section{Computational Infrastructure}

We use dual NVIDIA RTX A6000 GPUs to conduct experiments on training deep point-cloud autoencoders.  
Other applications are run on a MacBook Pro 14-inch (Nov 2023) equipped with an Apple M3 Max chip and 36 GB memory.

\section{Ablation Studies}
\label{sec:ablations}

Section~\ref{sec:bo-sphere} describes several modeling and parameter choices made for our implementation of Bayesian optimization.
We justify these choices here via ablation studies.
Each study is evaluated on the point-cloud interpolation (gradient flow) example of Section~\ref{sec:pci}.
For consistency, we also select $L = 100$ for all our ablation studies unless otherwise noted.

\begin{figure}[t]
  \centering
  % First subfigure
  \begin{subfigure}{0.49\textwidth}
    \centering
    \includegraphics[width=\linewidth]{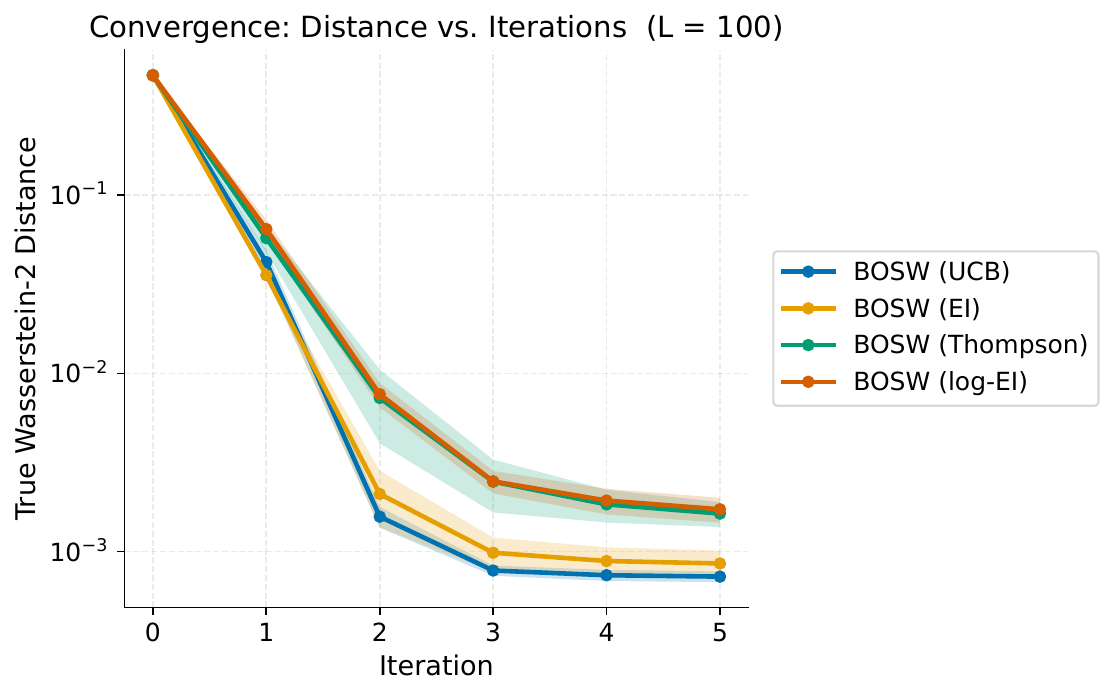}
    \caption{Convergence with increasing iterations.}
    \label{fig:ablation-acq-1}
  \end{subfigure}
  \hfill
  % Second subfigure
  \begin{subfigure}{0.49\textwidth}
    \centering
    \includegraphics[width=\linewidth]{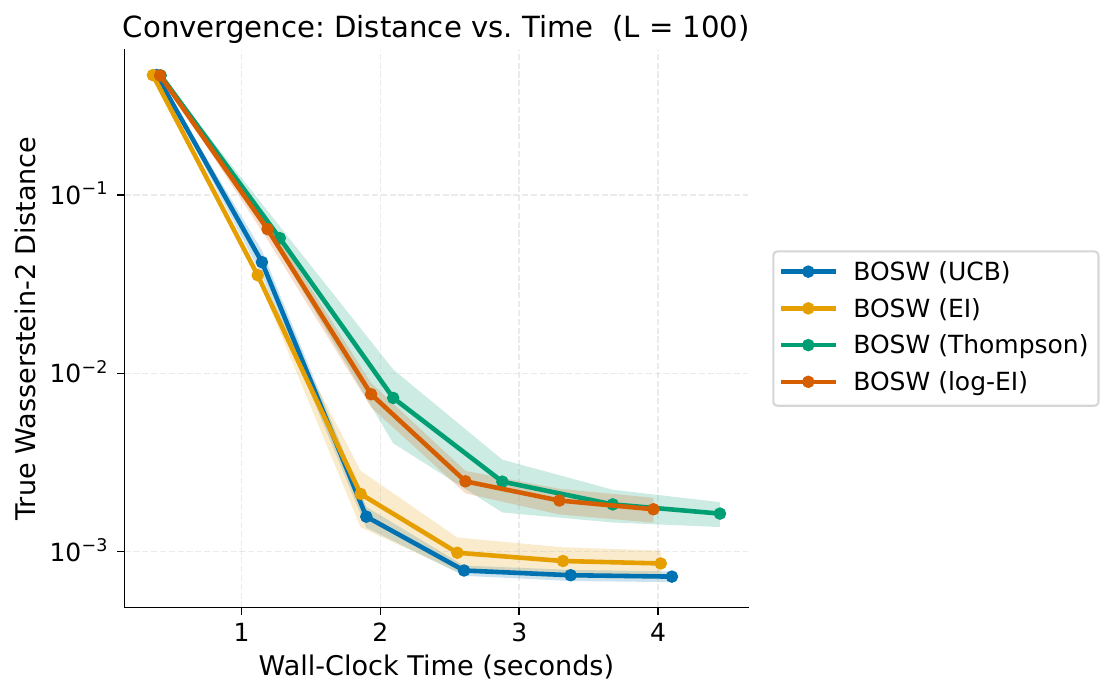}
    \caption{Convergence with increasing wall-clock time.}
    \label{fig:ablation-acq-2}
  \end{subfigure}
  
  \caption{Ablation study of different acquisition functions for Bayesian optimization, including upper confidence bound (UCB), expected improvement (EI), Thompson sampling, and log-EI.  UCB consistently performs best in terms of number of iterations and wall-clock time required to converge.}
  \label{fig:ablation-acq}
\end{figure}

First, we experiment with different possible acquisition functions for Bayesian optimization.
We test popular functions such as upper confidence bound (UCB), expected improvement (EI), Thompson sampling \citep{thompson1933likelihood}, as well as the recently-proposed log-EI \citep{10.5555/3666122.3667026}.
Figure~\ref{fig:ablation-acq} shows that UCB consistently outperforms other acquisition functions.
Thus, we select UCB for all of our examples in the paper.
Interestingly, we note that a recent work \citep{uuv} also found that LCB (identical to UCB, but for minimization rather than maximization) was the most efficient acquisition function choice, in a completely different application domain (design optimization).

\begin{figure}[t]
  \centering
  % First subfigure
  \begin{subfigure}{0.49\textwidth}
    \centering
    \includegraphics[width=\linewidth]{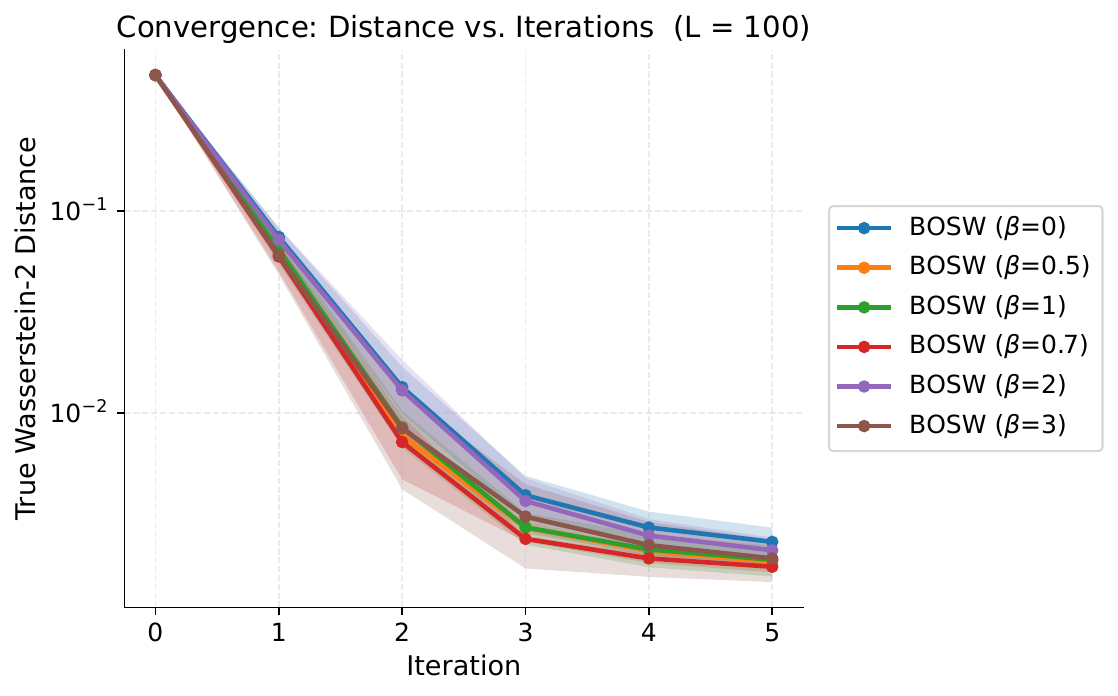}
    \caption{Convergence with increasing iterations.}
    \label{fig:ablation-beta-1}
  \end{subfigure}
  \hfill
  % Second subfigure
  \begin{subfigure}{0.49\textwidth}
    \centering
    \includegraphics[width=\linewidth]{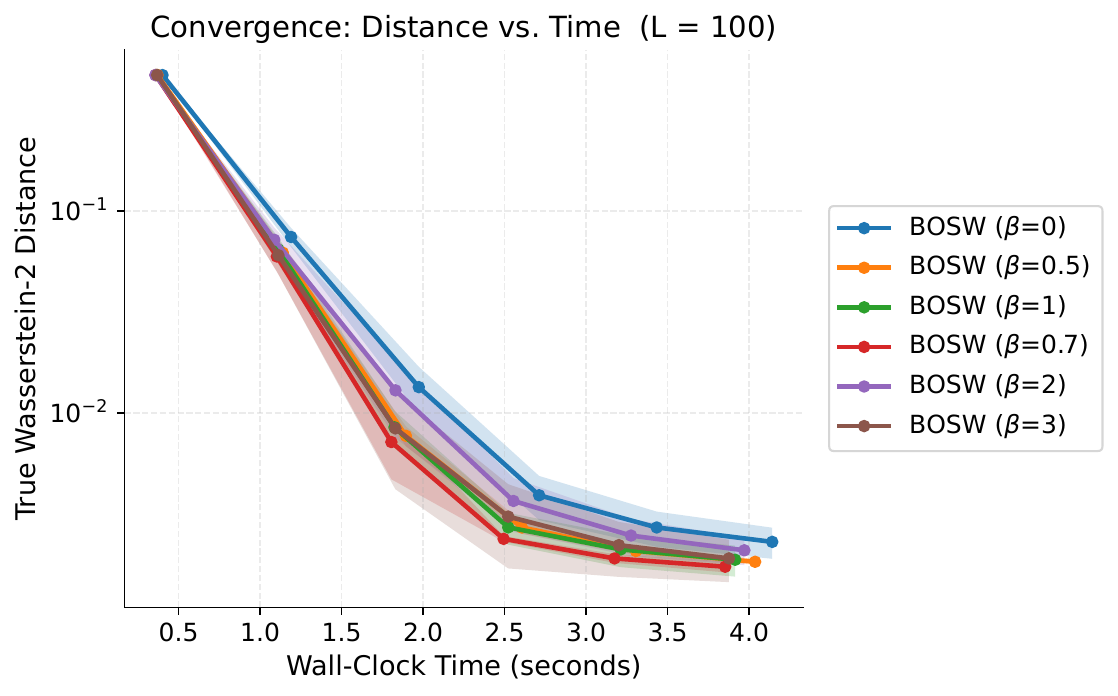}
    \caption{Convergence with increasing wall-clock time.}
    \label{fig:ablation-beta-2}
  \end{subfigure}
  
  \caption{Ablation study of different values of $\beta$ (the weight parameter for the UCB acquisition function).  Although the different values of $\beta$ yield similar results, $\beta = 0.7$ consistently demonstrates the best performance.}
  \label{fig:ablation-beta}
\end{figure}

Next, we consider different values of $\beta$, the parameter that tunes exploration vs.\ exploitation in the UCB acquisition function.
(Note that in our experiments, we use a constant $\beta$; adaptive $\beta$ schedules are certainly possible and may benefit bias (see Appendix~\ref{sec:theory}.)
As demonstrated in Figure \ref{fig:ablation-beta}, the constant value $\beta = 0.7$ that we select for all our experiments outperforms all other choices considered (both larger and smaller).

\begin{figure}[t]
  \centering
  % First subfigure
  \begin{subfigure}{0.49\textwidth}
    \centering
    \includegraphics[width=\linewidth]{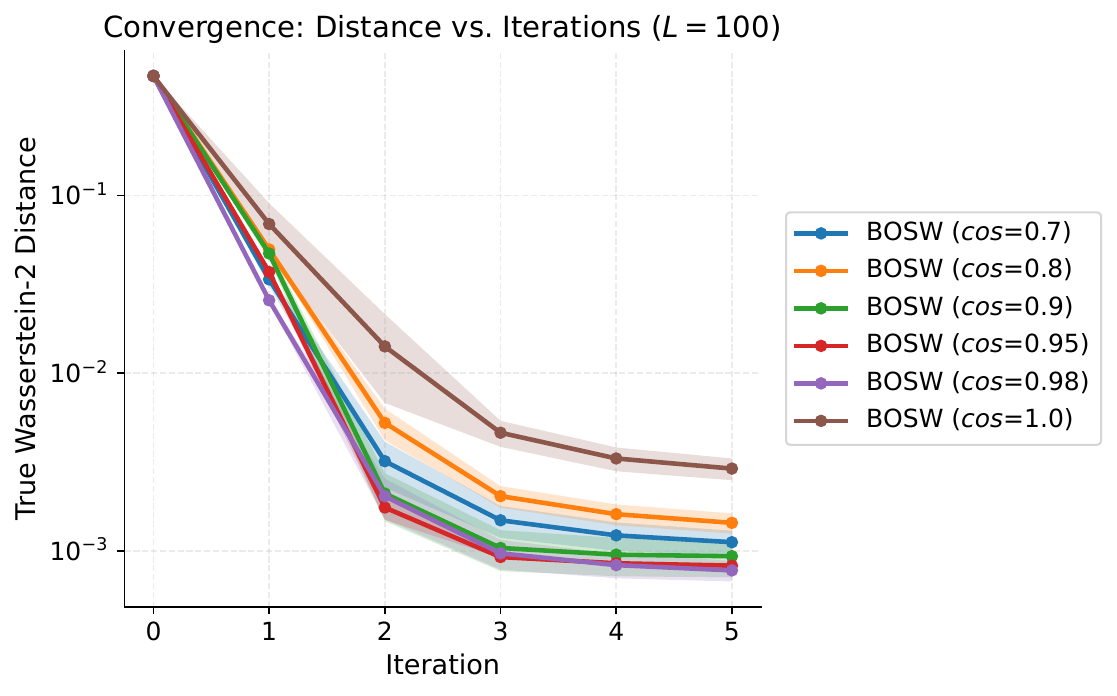}
    \caption{Convergence with increasing iterations.}
    \label{fig:ablation-cosine-1}
  \end{subfigure}
  \hfill
  % Second subfigure
  \begin{subfigure}{0.49\textwidth}
    \centering
    \includegraphics[width=\linewidth]{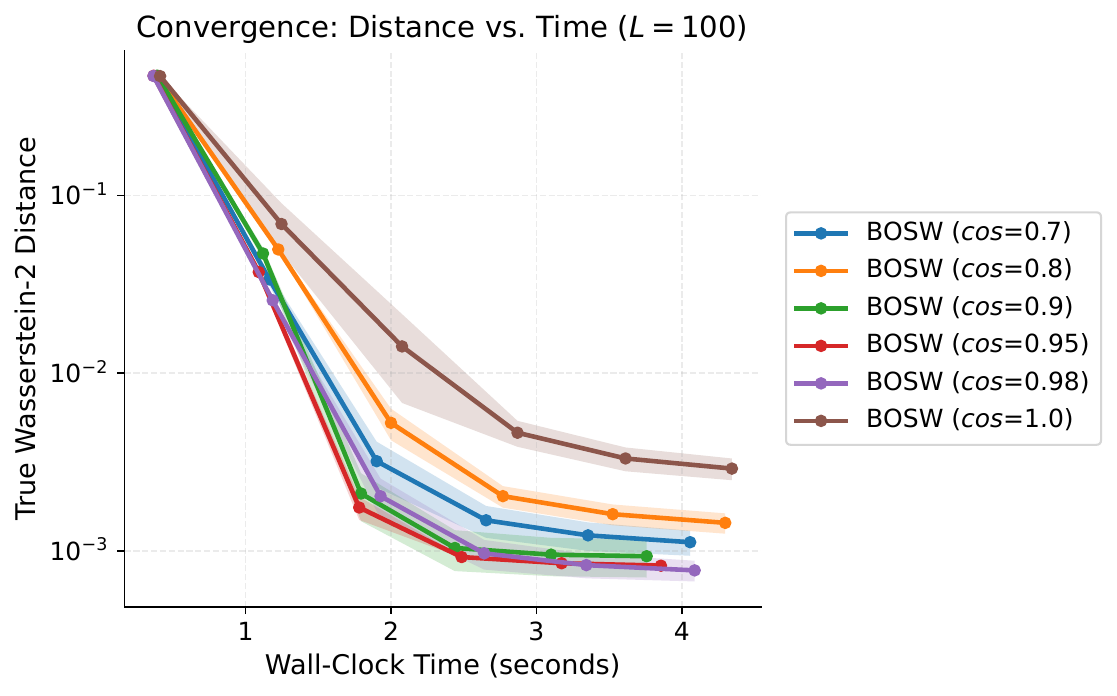}
    \caption{Convergence with increasing wall-clock time.}
    \label{fig:ablation-cosine-2}
  \end{subfigure}
  
  \caption{Ablation study of different values of the cosine-similarity cutoff threshold used in our method. The plots motivate our usage of $0.98$ for all our experiments.}
  \label{fig:ablation-cosine}
\end{figure}

We also explore different thresholds for the cosine-similarity cutoff (used to reject the selection of directions that are too close to each other).
We remark that this value should typically be chosen closer to 1 as $L$ increases, to facilitate large numbers of slices becoming dense on the unit hypersphere.
For $L = 100$, though (as generally used in our experiments), Figure~\ref{fig:ablation-cosine} shows that our choice of $0.98$ is near-optimal; while sometimes a slightly lower value ($0.9$ or $0.95$) helps achieve a slightly lower error at a particular time, a value of $0.98$ results in the lowest error after sufficient iterations or time.
An annealing scheme for a dynamic threshold value may unlock marginally better performance (particularly for large $L$), but we feel this would be a very minor optimization, especially since $L$ is usually kept relatively small to maintain computational efficiency of SW.

\begin{figure}[t]
    \centering
    \includegraphics[width=0.5\linewidth]{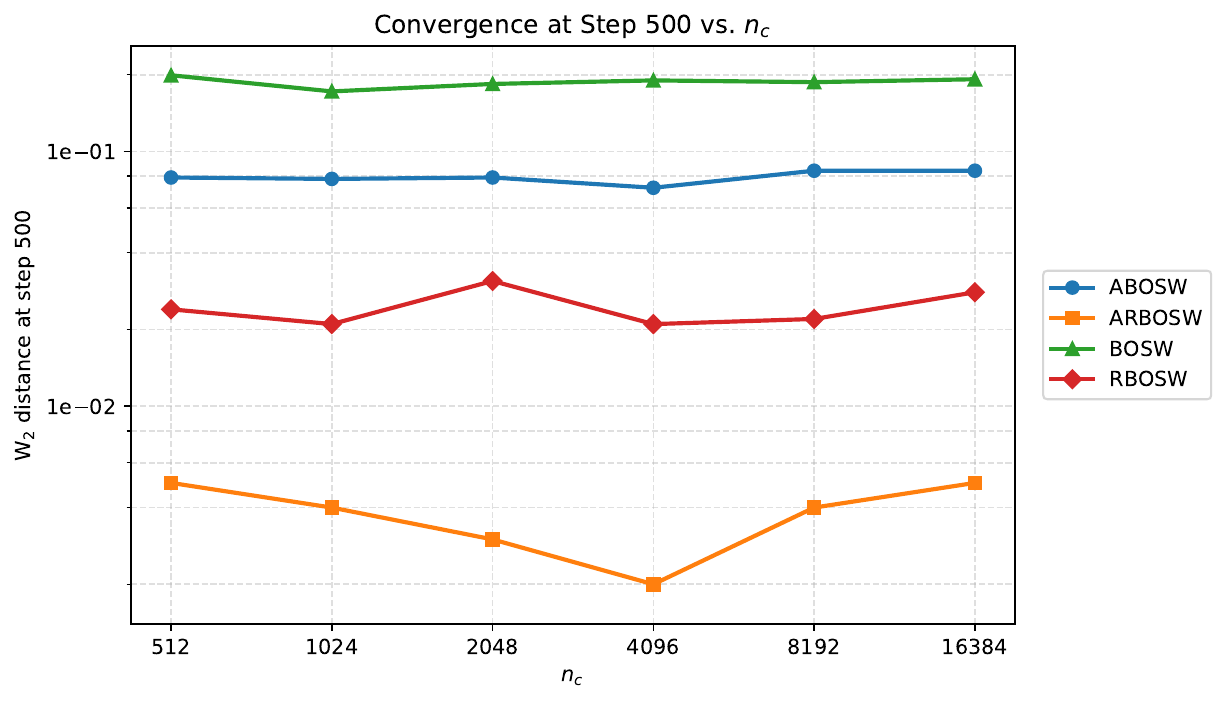}
    \caption{Effect of varying $n_c$ on the convergence of our BO-based methods.  For most methods, $n_c=4096$ is optimal, except for BOSW, which attains minimal error after 500 steps using $n_c = 1024$.  Note that the vertical axis is scaled logarithmically to make the data points easier to distinguish.}
    \label{fig:ablation-nc}
\end{figure}

Finally, we consider different values of $n_c$, the number of candidate directions considered on any iteration of our core BO routine.
We chose $n_c = 4096$ to copy the number of candidate directions used in the QSW methods of \citet{quasiSW}.
Interestingly, as Figure~\ref{fig:ablation-nc} shows, this value turns out to be optimal for all methods except BOSW (where 1024 is optimal).
Nonetheless, we found that values from 512 to 16384 did not significantly change the error in our methods after 500 steps, despite the linear-time computational complexity associated with $n_c$.
Thus, in practice, it may be appropriate to use smaller values of $n_c$, accepting a modest change in accuracy for a potentially substantial change in performance.
We note that, as expected, for sufficiently small values of $n_c$, no method converged well (neither QSW methods nor our BO-based methods).

\section{Theoretical Performance of Bayesian Optimization for Sliced Wasserstein}
\label{sec:theory}

Although our paper is primarily intended to provide numerical evidence of the efficacy of using Bayesian optimization for sliced Wasserstein distance computations, we briefly remark on the possibility of theoretical performance bounds.

It is known that under i.i.d.\ MC sampling and sufficient regularity assumptions, SW converges at a rate of $O(L^{-1/2})$.
By the triangle inequality, we can relate the error of our BO methods to the error of MC SW:
\begin{align*}
&\left|SW_L^p(\hat{\Theta}_T) - SW_p^p(\mu,\nu)\right| \leq
\underbrace{
    \left|SW_L^p(\hat{\Theta}_T) - SW_L^p(\Theta_\text{MC})\right|
}_{\mathclap{\text{BO-MC Error}}}
+ 
\underbrace{
    \left|SW_L^p(\Theta_\text{MC}) - SW_p^p(\mu,\nu)\right|
}_{\mathclap{\text{MC-SW Error}}} .
\end{align*}
The latter term is bounded above by $O(L^{-1/2})$, while bounding the former term remains an open question.
The primary reason bounding that term is problematic is that BOSW may \textit{not} perform as well as MC SW in the limit, because BOSW is not generally an unbiased estimator.
For instance, with our UCB acquisition function as used in our paper,
$$
\alpha_t(\theta) = \mu_{t-1}(\theta) + \beta \sigma_{t-1} (\theta),
$$
we take a constant $\beta = 0.7$.
Thus, $\alpha_t$ will always be encouraged to select $\theta$ yielding high $f(\theta; \mu, \nu)$.
In the limit, this suggests that
$$
\frac{1}{L} \sum_{i=1}^{L} f(\hat \theta_i) \to \sup_{\theta \in S^{d-1}} f(\theta) > \mathbb{E}_{\theta \sim U}[f(\theta)] .
$$
This is a strict inequality assuming $f$ is non-constant on $S^{d-1}$; and when sampling $\theta$ uniformly (MC), as mentioned above, $\mathbb{E}_{\theta \sim U}[f(\theta)] = 0$.
This rough sketch demonstrates that BOSW, at least with UCB with a constant $\beta$, will not converge in the limit---despite its attractive numerical performance (demonstrated in our experiments in the main body) during early iterations.
Bounding the performance of BOSW for finite $L$ is an interesting open question.
It may be possible to approach this question using the concepts of information gain and regret minimization; see, e.g., \citet{vakili2021information}.
This would be impacted by our choice of using an angular RBF kernel.

Despite gaps in the current theory, we note that we can use an annealing schedule for $\beta$ to drive our GP-UCB sampler towards eventual uniform coverage of the sphere.
In spirit, this follows the idea of using BOSW for early iterations, then switching to MC or other variants for later iterations---an idea that could harmonize the benefits of fast early convergence with BO-based methods with the benefits of uniform coverage of MC samplers.
For instance, if we define
$$
\hat \alpha_t(\theta) = \epsilon_t \alpha_t(\theta) + (1-\epsilon_t)\frac{1}{|S^{d-1}|} ,
$$
we can choose $\epsilon_t = t^{-\gamma}$ ($0 < \gamma < 1$).
Then as $t \to \infty$, $\epsilon_t \to 0$, but $\sum_t \epsilon_t = \infty$, so informally speaking\footnote{The lemma assumes each ``event'' (direction selection) is independent, which is not the case, by design.  Nonetheless, this assumption \textit{would} hold in the limit of $\epsilon_t \to 0$.  Generalizations of the Borel-Cantelli lemma that do not require strict independence have been studied \citep{CHANDRA2008390,biró2021weakindependenceeventsconverse}, and it is possible that those may apply here.  For the present work, though, we only characterize our arguments as informal.}, the second Borel–Cantelli lemma would suggest that every measurable region of the hypersphere is sampled infinitely often with probability 1.
It follows that the bias will match that of uniform MC sampling, which is 0 in the limit; and under this scheme, the convergence rate would reach $O(L^{-1/2})$ asymptotically (whether it is faster or slower during initial iterations).

\section{Use of Large Language Models}

Large language models (LLMs) were used in the preparation of this paper.
Specifically, Perplexity was prompted to act as a peer reviewer for our draft.
In this context, the model identified several recent references for us to discuss in our related work that we previously had not included; and the model reminded us to perform a few of the ablation studies seen in Appendix~\ref{sec:ablations}.

\end{document}